\documentclass[letterpaper, 10 pt, conference]{ieeeconf}  
\pdfoutput=1

\makeatletter
\let\NAT@parse\undefined
\makeatother

\IEEEoverridecommandlockouts                              

\usepackage{amsmath} 

\usepackage{graphicx}
\usepackage{caption}
\usepackage{subcaption}
\usepackage{tabularx}
\usepackage{bm}
\usepackage{amsmath}
\usepackage{amsfonts}
\usepackage{hyperref}
\usepackage{color}

\usepackage[ruled,vlined]{algorithm2e}

\DeclareMathOperator*{\argmax}{arg\!\max}

\title{\LARGE \bf
4D Crop Monitoring: Spatio-Temporal Reconstruction for Agriculture
}

\author{Jing Dong, John Gary Burnham, Byron Boots, Glen C. Rains, Frank Dellaert
\thanks{
J. Dong, B. Boots and F. Dellaert are with College of Computing, Georgia Institute of Technology, USA. 
{\tt\small jdong@gatech.edu, \{bboots,frank.dellaert\}@cc.gatech.edu}.
J. G. Burnham and G. C. Rains are with the Department of Entomology, University of Georgia, USA. 
{\tt\small \{burnhamj,grains\}@uga.edu}}%
}

\begin{document}

\maketitle
\thispagestyle{empty}
\pagestyle{empty}

\begin{abstract}
Autonomous crop monitoring at high spatial and temporal resolution is a critical problem in precision agriculture. 
While Structure from Motion and Multi-View Stereo algorithms can finely reconstruct the 3D structure of a field with low-cost image sensors, these algorithms fail to capture the dynamic nature of continuously growing crops.
In this paper we propose a \emph{4D reconstruction} approach to crop monitoring, which employs a spatio-temporal model of dynamic scenes that is useful for precision agriculture applications.
Additionally, we provide a robust data association algorithm to address the problem of large appearance changes due to scenes being viewed from different angles at different points in time, which is critical to achieving 4D reconstruction.
Finally, we collected a high quality dataset with ground truth statistics to evaluate the performance of our method.
We demonstrate that our 4D reconstruction approach provides models that are qualitatively correct with respect to visual appearance and quantitatively accurate when measured against the ground truth geometric properties of the monitored crops.
\end{abstract}

\section{Introduction \& Related Work}
Automated crop monitoring is a key problem in precision agriculture, used to maximize crop yield while minimizing cost and environmental impact. 
Traditional crop monitoring techniques are based on measurements by human operators, which is both expensive and labor intensive.
Early work on automated crop monitoring mainly relies on satellite imagery~\cite{Rembold13using}, which is expensive and lacks sufficient resolution in both space and time.
Recently, crop monitoring with Unmanned Aerial Vehicles (UAVs)~\cite{Bryson10airborne,Anthony14iros,Das15devices,Zainuddin16cspa,Sarkar16towards} and ground vehicles~\cite{Hague06automated,Lalonde06automatic,Singh10comprehensive,Nuske14automated} has garnered interest from both agricultural and robotics communities due to the abilities of these systems to  gather large quantities of data with high spatial and temporal resolution.

Computer vision is a powerful tool for monitoring the crops and estimating yields with low-cost image sensors~\cite{Hague06automated,Nuske14automated,Font15vineyard,Sa16deepfruits}. However, the  majority of this work only utilizes 2D information in individual images, failing to recover the 3D geometric information from sequences of images.
Structure from Motion (SfM)~\cite{Agarwal09iccv} is a mature discipline within the computer vision community that enables the recovery of 3D geometric information from image. 
When combined with Multi-View Stereo (MVS) approaches~\cite{Furukawa10pami}, these methods can be used to obtain dense, fine-grained  3D reconstructions. 
The major barrier to the direct use of these methods for crop monitoring is that traditional SfM and MVS methods only work for \emph{static scenes}, which cannot solve 3D reconstruction problems with \emph{dynamic growing} crops.

\begin{figure}[!t]
\centering
{\includegraphics[width=0.98\columnwidth]{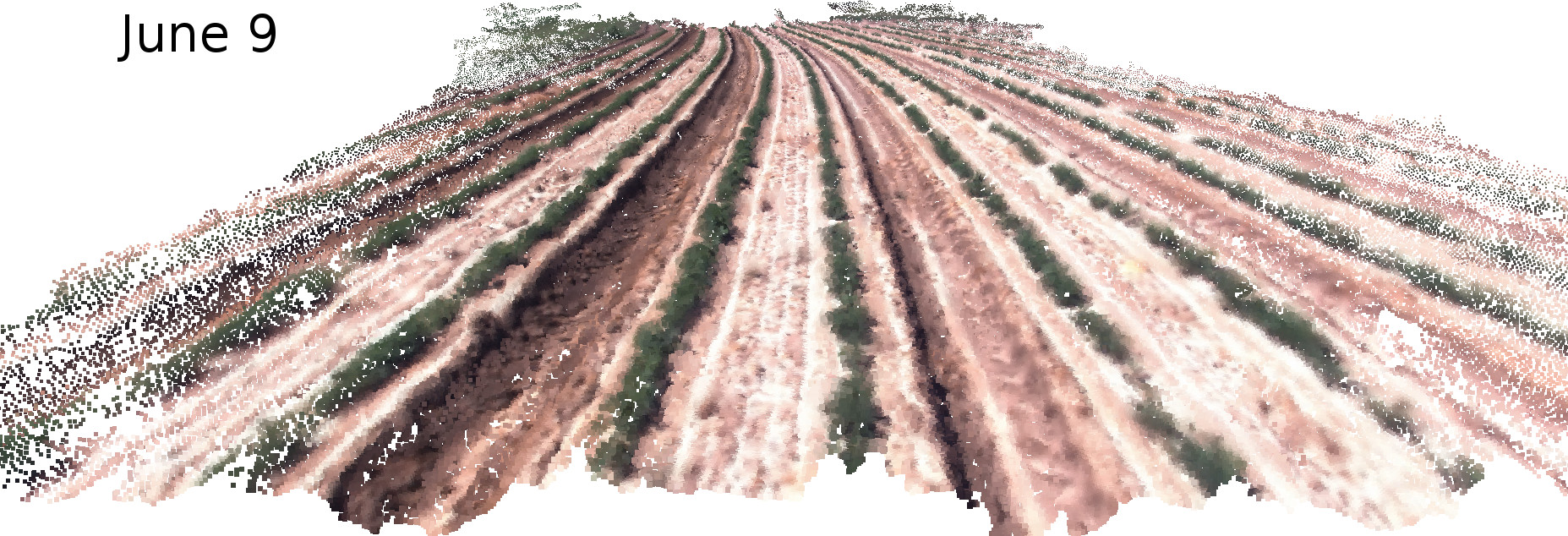}}\\
\vspace*{3mm}
{\includegraphics[width=0.98\columnwidth]{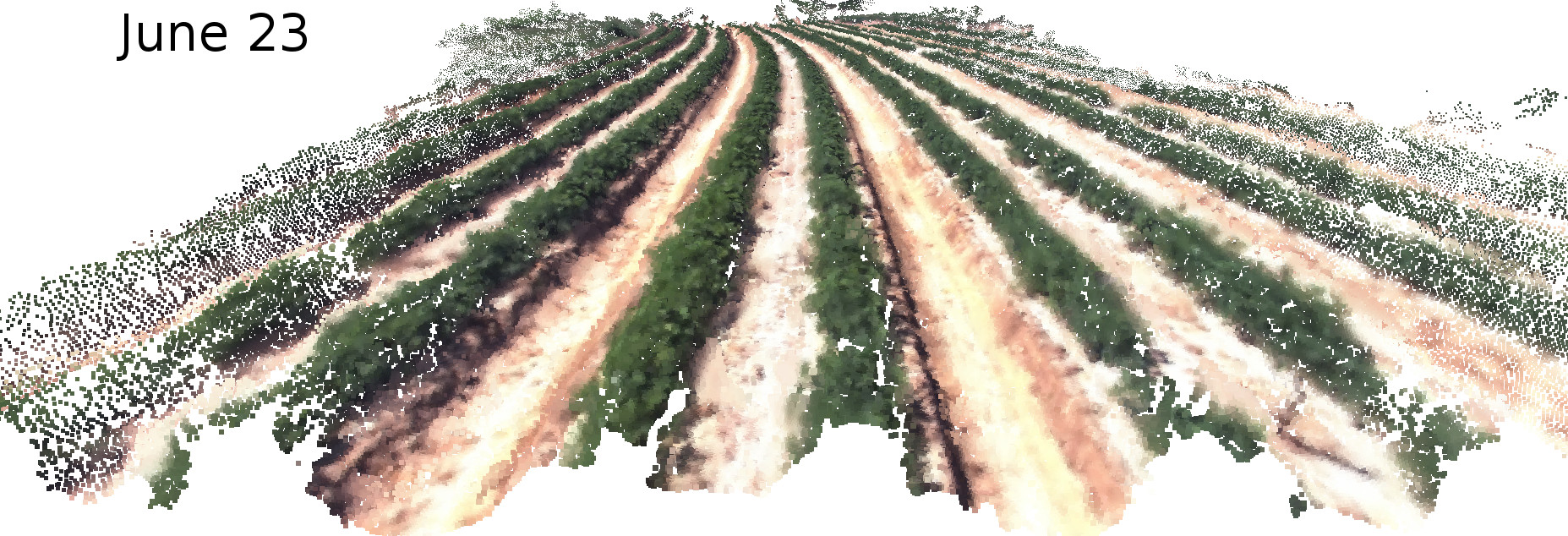}}\\
\vspace*{3mm}
{\includegraphics[width=0.98\columnwidth]{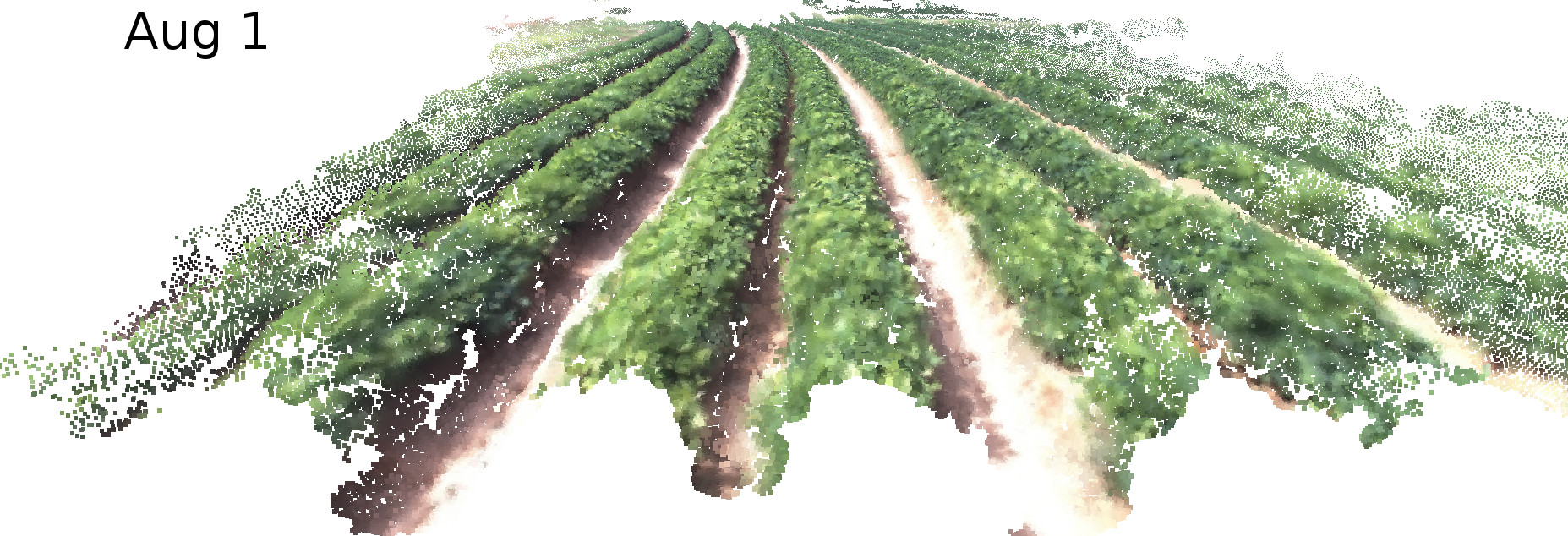}}
\caption{Reconstructed 4D model of a peanut field by our approach. Each time slice of the 4D model is a dense reconstructed point cloud.}
\label{fig:4d_results_1}
\end{figure}

Change detection in both 2D images~\cite{Sakurada13cvpr,Alcantarilla16rss}, 3D point clouds~\cite{Xiao15isprs}, and volumetric reconstruction~\cite{Pollard07cvpr,Ulusoy14eccv}, have been extensively studied. 
Taneja et al.~\cite{Taneja11iccv} generates a 3D reconstruction of changed parts from 2D images, and has applied the approach to city-scale change detection and 3D reconstruction~\cite{Taneja13cvpr,Taneja15pami}.
3D change detection has also been applied to urban construction site monitoring~\cite{Golparvar11iccvws}. However, this type of scene change detection, which focuses on local changes, does not apply to precision agriculture applications where crops are \emph{constantly} changing over time.

Related previous work includes data association approaches that build visual correspondences between scenes with appearance variations. 
Data association is a key element of 3D reconstruction: by identifying geometric correspondences between scenes with appearance changes (e.g. due to changing viewpoint), the scene can be  reconstructed while minimizing misalignment.
Recently, data association has been applied to even more difficult problems. For example, Griffith et al. studied the alignment of natural scene images with significant seasonal changes~\cite{Griffith15rssws,Griffith16bmvc}.
Localization and place recognition are applications that require data association approaches that are highly robust to illumination and seasonal appearance changes~\cite{Milford12icranew,Naseer15ecmp}.
A more comprehensive survey of visual place recognition and localization can be found in~\cite{Lowry16tro}.
Particularly, Beall et al. build a spatio-temporal map, which helps to improve robustness of localization across different seasons~\cite{Beall14ppniv}.
The difficulty with applying these existing methods to field reconstruction is that they are usually designed for autonomous driving applications, in which the camera has few view angle variations. As a result, these approaches cannot solve the large-baseline data association problems that are prevalent in field reconstruction due to the large view angle variations typically encountered by vehicles traversing a field.

The computer vision community has worked on time-lapse reconstruction of dynamic scenes for years, but most existing approaches do not obviously apply to crop monitoring applications.
Martin et al.~\cite{Martin15siggraph,Martin15iccv} synthesizes smooth time-lapse videos from images collected on the Internet, but this work is limited to 2D results, without any 3D geometric information. 
Early work on 4D reconstruction includes \cite{Schindler07cvpr,Schindler10cvpr}, which build city-scale 3D reconstructions via temporal inference from historical images.
Further work includes~\cite{Matzen14eccv}, which offers better scalability and granularity.
A probabilistic volumetric 4D representation of the environment was proposed by~\cite{Ulusoy13iccv}, and then used in actual 4D reconstructions enabled by 3D change detection~\cite{Ulusoy14eccv}.
The major issue with most existing approaches is that they assume each geometric entity keeps nearly-constant appearance for the temporal duration, which is not the case in crop monitoring since crops are changing continuously.

In this paper we address the problem of time-lapse 3D reconstruction with \emph{dynamic scenes} to model \emph{continuously growing} crops. 
We call the 3D reconstruction problem with temporal information \emph{4D reconstruction}.
The output of 4D reconstruction is a set of 3D entities (point, mesh, etc.), associated with a particular time or range of times. An example is shown in Fig.~\ref{fig:4d_results_1}.
A 4D model contains all of the information of a 3D model, e.g. canopy size, height, leaf color, etc., but also contains additional temporal information, e.g. growth rate and leaf color transition.

We also collected a field dataset using a ground vehicle equipped with various sensors, which we will make publicly available. To our knowledge, this will be first freely available dataset that contains large quantities of spatio-temporal data for robotics applications targeting precision agriculture.

Our paper contains three main contributions:
\begin{itemize}
\item We propose an approach for 4D reconstruction for fields with continuously changing scenes, mainly targeting crop monitoring applications.
\item We propose a robust data association algorithm for images with highly duplicated structures and significant appearance changes.
\item We collect a ground vehicle field dataset with ground truth crop statistics for evaluating 4D reconstruction and crop monitoring algorithms.
\end{itemize}

\section{Method} \label{sec:method}

\begin{figure}[!t]
\centering
{\includegraphics[width=0.5\columnwidth]{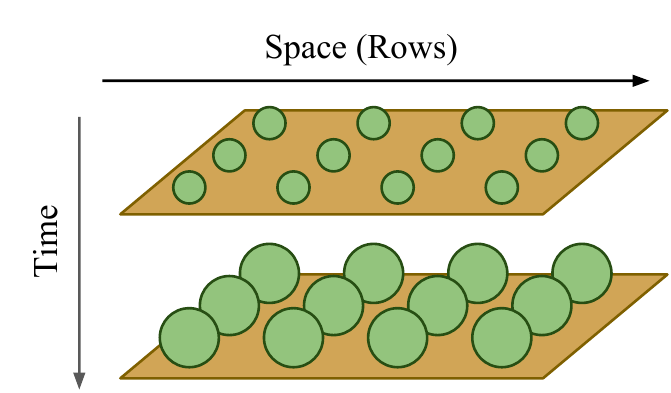}}
\caption{The field 4D model. The field contains multiple rows, and there are multiple time sessions of the field.}
\label{fig:field}
\end{figure}

We begin by stating several assumptions related to crop monitoring, before specifying the details of our 4D reconstruction algorithm. 
\begin{itemize}
\item The scene is \emph{static} during each data collection session.
\item The field may contain multiple \emph{rows}.
\end{itemize}
The first assumption is acceptable because we only focus on modelling crops and ignore other dynamic objects like humans, and the crop growth is too slow to be noticeable during a single collection session. 
The second assumption is based on the geometric structure of most fields.
The 4D field model reflecting these two assumptions is illustrated in Fig.~\ref{fig:field}.

Our proposed system has three parts. 
\begin{itemize}
\item A multi-sensor Simultaneous Localization and Mapping (SLAM) pipeline, used to compute camera poses and field structure for a \emph{single} row in a \emph{single} session. 
\item A data association approach to build visual correspondences between different rows and sessions. 
\item A optimization-based approach to build the full 4D reconstruction across all rows and all sessions.
\end{itemize}
To generate the 4D reconstruction of the entire field, we first compute 3D reconstruction results for each row at each time session, by running multi-sensor SLAM independently.  
Next we use the data association approach to match images from different rows and sessions, building a joint factor graph that connects the individual SLAM results. 
Finally we optimize the resultant joint factor graph to generate the full 4D results.

To clarify notation, we assign the superscript of each symbol to the row index and time session index in the remaining article, unless otherwise mentioned. 
The superscript $\langle t_i, r_j \rangle$ indicates the variable is associated to row $r_j$ of the field at time session $t_i$.

\subsection{Multi-Sensor SLAM}

\begin{figure}[!t]
\centering
{\includegraphics[width=1.0\columnwidth]{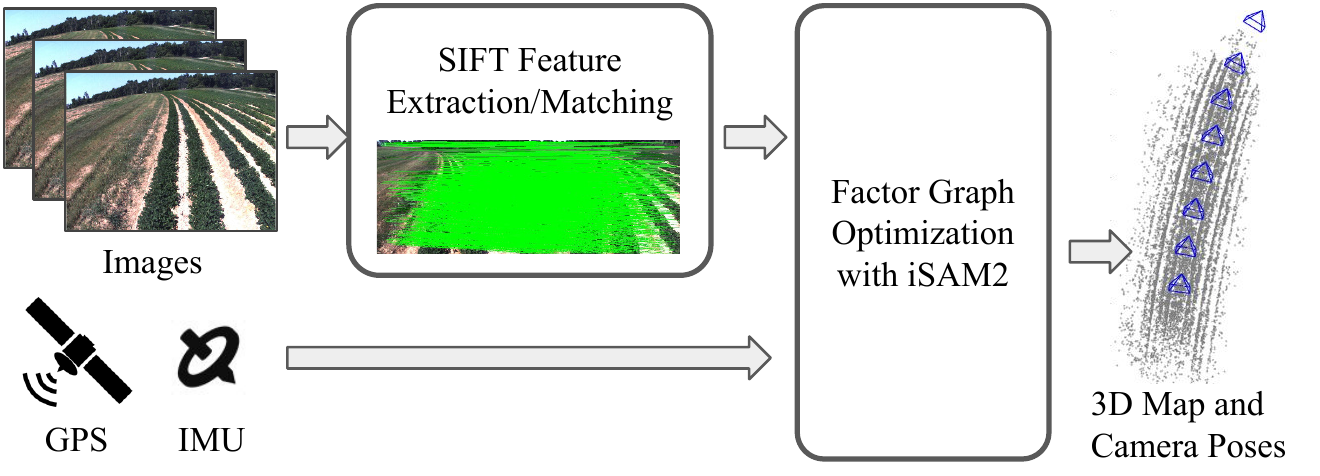}}
\caption{Overview of multi-sensor SLAM system.}
\label{fig:slam_pipeline}
\end{figure}

The SLAM pipeline used in this work has two parts, illustrated in Fig.~\ref{fig:slam_pipeline}.

The first part of the SLAM system is a front-end module to process images for visual landmarks. 
SIFT~\cite{Lowe04ijcv} features are extracted from each image, and SIFT descriptor pairs in nearby image pairs are matched by the approximate nearest neighbor library FLANN~\cite{Muja14pami}.
The matches are further filtered by 8-point RANSAC~\cite{Hartley04book} to reject outliers.
Finally a single visual landmark is accepted if there are more than 6 images that have corresponding features matched to the same landmark.

The second part of the SLAM system is a back-end module for estimating camera states and landmarks using visual landmark information from the front-end and other sensor inputs.
Since the goal of the multi-sensor SLAM system is to reconstruct a \emph{single} row during a \emph{single} data collection session, the back-end module of the SLAM system estimates a set of $N$ camera states $X^{\langle t_i, r_j \rangle} = \{\bm{x}^{\langle t_i, r_j \rangle}_0, ..., \bm{x}^{\langle t_i, r_j \rangle}_{N-1}\}$ at row $r_j$ and time $t_i$, given visual landmark measurements from the front-end module, and other sensor measurements, including an Inertial Measurement Unit (IMU) and GPS.
For each camera state, we estimate $\bm{x}_j = \{\mathbf{R}_j, \mathbf{t}_j, \bm{v}_j, \bm{\omega}_j, \mathbf{b}_j\}$, which includes 
camera rotation $\mathbf{R}_j$, camera translation $\mathbf{t}_j$, translational velocity $\bm{v}_j$, angular rotation rate $\bm{\omega}_j$, and the IMU sensor bias $\mathbf{b}_j$.

The SLAM problem is formulated on a factor graph~\cite{Dellaert06ijrr} where
the joint probability distribution of estimated variables $X^{\langle t_i, r_j \rangle}$ given measurements $Z^{\langle t_i, r_j \rangle}$ is factorized and represented as the product 
\begin{equation}
p(X^{\langle t_i, r_j \rangle} | Z^{\langle t_i, r_j \rangle}) \propto \prod^{K}_{k=1} \phi(X^{\langle t_i, r_j \rangle}_k),
\end{equation}
where $K$ is the total number of factors, $X^{\langle t_i, r_j \rangle}_k$ is the set of variables the $k$th factor involved, and $\phi$ is the factor in the graph which is proportional to measurement likelihood $l(X^{\langle t_i, r_j \rangle}_k; z^{\langle t_i, r_j \rangle}_k)$, given $k$th measurement $z^{\langle t_i, r_j \rangle}_k \in Z^{\langle t_i, r_j \rangle}$.
The states can then be computed by Maximum a Posteriori (MAP) estimation
\begin{equation}
\hat{X}^{\langle t_i, r_j \rangle} = \argmax_{X^{\langle t_i, r_j \rangle}} p(X^{\langle t_i, r_j \rangle} | Z^{\langle t_i, r_j \rangle}).
\end{equation}

\begin{figure}[!t]
\centering
{\includegraphics[width=0.6\columnwidth]{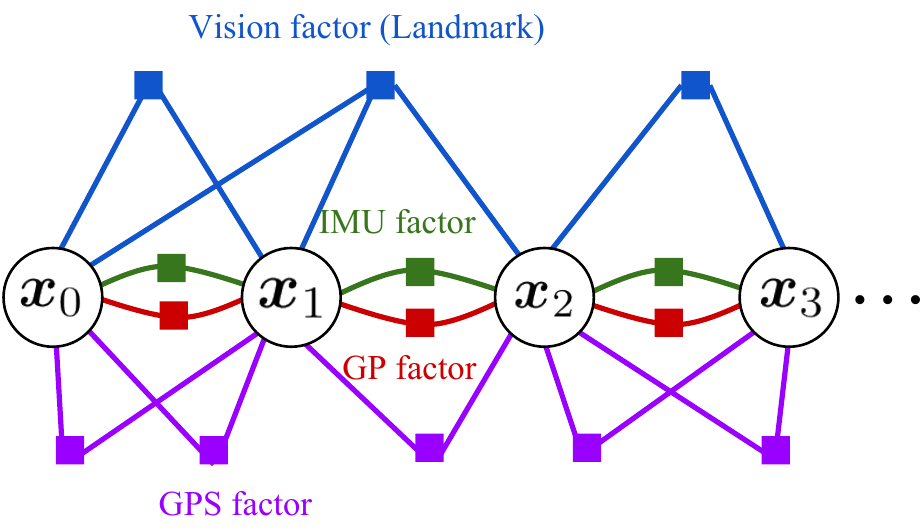}}
\caption{Factor graph of multi-sensor SLAM.}
\label{fig:slam_factor_graph}
\end{figure}

The details of the factor graph is shown in Fig.~\ref{fig:slam_factor_graph}.
We use smart factors~\cite{Carlone14icra} for visual landmarks, to reduce memory storage by avoiding the explicit estimation of landmark variables.
Outlier rejection 
is used to reject landmarks with re-projection error larger than 10 pixels. 
IMU measurements are incorporated into the factor graph by preintegrated IMU factors~\cite{Forster15rss}.
GPS measurements are not synchronized with images (detailed in Sec.~\ref{sec:dataset}), so we use 
a continues-time SLAM approach which formulates the SLAM problem as a Gaussian Process (GP)~\cite{Anderson15iros}, so the GPS measurements are easily incorporated into factor graph as interpolated binary factors (magenta factors in Fig.~\ref{fig:slam_factor_graph}), with GP prior factors (red factors in Fig.~\ref{fig:slam_factor_graph}). Details about continuous-time SLAM as a GP can be found in \cite{Barfoot14rss,Anderson15iros,Yan15isrr}.

We optimize the factor graph by iSAM2~\cite{Kaess12ijrr}.
Once camera states are estimated, $M$ landmarks $L^{\langle t_i, r_j \rangle} = \{l^{\langle t_i, r_j \rangle}_0, ..., l^{\langle t_i, r_j \rangle}_{M-1}\}$ are triangulated by known camera poses.

\subsection{Robust Data Association over Time and Large Baseline}

The second key element of our approach is robust data association.
Data association is a key technique to get reconstruction results of more than a single row at a single time; however, the data association problem between different rows or times is difficult, since there are significant appearance changes due to illumination, weather or view point changes.
The problem is even more difficult in crop monitoring due to \emph{measurement aliasing}~\cite{Indelman16csm}: fields contain highly periodic structures with little visual difference between plants (see Fig.~\ref{fig:4d_results_1}).
As a result, data association problems between different rows and times is nearly impossible to solve by image-only approaches.

Rather than trying to build an image-only approach, 
we use single row reconstruction results output by SLAM as a starting point for data association across rows and time. 
The SLAM results provide camera pose and field structure information from all of the sensors (not just images), which helps us to improve the robustness of data association. 

\begin{figure}[!t]
\centering
{\includegraphics[width=0.7\columnwidth]{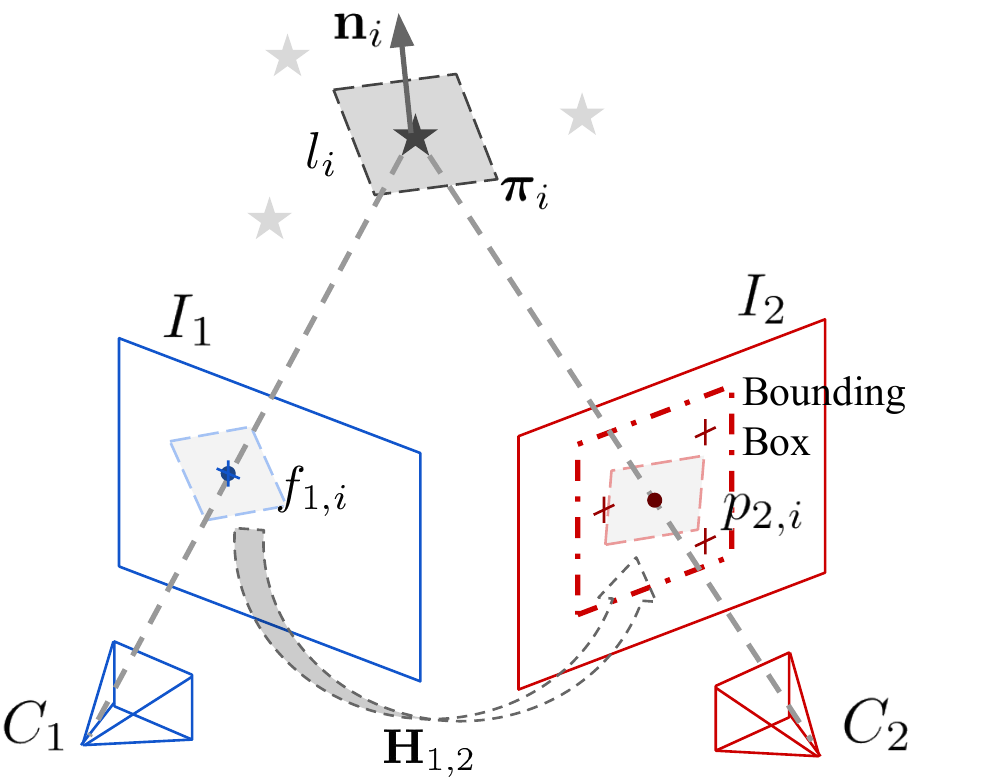}}
\caption{Diagram of robust data association.}
\label{fig:data_association_diag}
\end{figure}

Specifically, the data association problem involves finding visual correspondences (matches between SIFT feature points) between two images, $I_1$ and $I_2$, which are taken by camera $C_1$ and $C_2$ respectively, as shown in Fig.~\ref{fig:data_association_diag}. 
Cameras $C_1$ and $C_2$ may come from the same or different rows, during the same or different time sessions.
Each camera $C = \{\mathbf{K}, \mathbf{R}, \mathbf{t}\}$ contains the intrinsic calibration $\mathbf{K}$ which is known, and camera pose $\{\mathbf{R}, \mathbf{t}\}$ estimated by SLAM.

We combine two methods to build a data association using prior information from SLAM, \emph{back projection bounded search} and \emph{homography image warping}. The two methods are detailed here.


\subsubsection*{Back Projection Bounded Search}

The basic idea of back projection bounded search is to reduce number of possible outliers by limiting the search range while seeking visual correspondences.
Assume $L_1$ is the set of all estimated landmarks visible in $C_1$, and each landmark in $L_1$ has corresponding feature points in $I_1$. 
For each $l_i \in L_1$, the linked feature point $f_{1,i} \in I_1$ might have a corresponding matched point at $p_{2,i} \in I_2$, which is the back-projected point of $l_i$ on $I_2$, if $C_1$, $C_2$ and $l_i$ are accurately estimated and $l_i$ does not change its appearance.
With estimation errors, we define a relaxed search area as a bounding box centered at $p_{2,i}$ on $I_2$, as shown in Fig.~\ref{fig:data_association_diag}, to search for the corresponding feature point for $f_{1,i}$. 
This significantly limits the search area to match $f_{1,i}$ and reject many possible  outliers, compared with searching the whole of $I_2$.

\begin{figure}
\centering
\begin{subfigure}[b]{0.1\textwidth}
\centering
\includegraphics[width=1\linewidth]{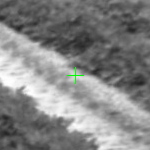}\\
\vspace*{2mm}
\includegraphics[width=1\linewidth]{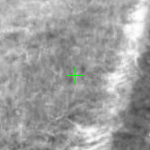}
\caption{$I_1$}
\end{subfigure}
\hspace{6mm}
\begin{subfigure}[b]{0.1\textwidth}
\centering
\includegraphics[width=1\linewidth]{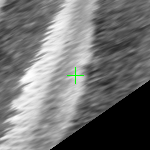}\\
\vspace*{2mm}
\includegraphics[width=1\linewidth]{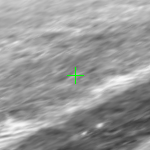}
\caption{$I'_1$}
\end{subfigure}
\hspace{6mm}
\begin{subfigure}[b]{0.1\textwidth}
\centering
\includegraphics[width=1\linewidth]{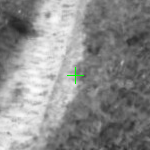}\\
\vspace*{2mm}
\includegraphics[width=1\linewidth]{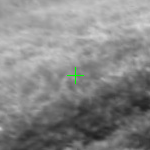}
\caption{$I_2$}
\end{subfigure}
\protect\caption{
(a) original $I_1$, (b) warped image $I'_1$, and (c) original $I_2$.
Patch center with green cross is feature point $f_{1,i}$ on (a), $f'_{1,i}$ on (b), and back project point $p_{2,i}$ on (c).
\label{fig:warped_patches}}
\end{figure}

\subsubsection*{Homography Image Warping}

Although the back projection bounded search rejects the majority of outliers, data association is still difficult when the viewing angle changes: the object's appearance may change significantly with large baselines, causing the search for a match in object's appearance from $I_1$ to $I_2$ based on SIFT descriptors, which also change with appearance. This is the major challenge for data association across images collected in different rows.

To combat this problem, we use a homography based method to eliminate appearance variations in $l_i$ due to viewpoint changes.
We assume that $l_i$ lies on a local \emph{plane} $\bm{\pi}_i$. If this assumption is satisfied, $\bm{\pi}_i$ induces a homography $\mathbf{H}_{1,2}$ from $I_1$ to $I_2$~\cite[p.327]{Hartley04book}
\begin{equation}
\mathbf{H}_{1,2} = \mathbf{K}_2 \big( \mathbf{R}_{1,2} - \frac{\mathbf{t}_{1,2}\mathbf{n}_i^{\top}}{d} \big) \mathbf{K}_1^{-1}
\label{eq:H_12}
\end{equation}
where $\{\mathbf{R}_{1,2}, \mathbf{t}_{1,2}\}$ define the relative pose from $C_1$ to $C_2$, $\mathbf{n}_i$ is the normal vector of $\bm{\pi}_i$, and $d$ is the distance from $C_1$ to $\bm{\pi}_i$. 
We use $\mathbf{H}_{1,2}$ warp $I_1$ to get $I'_1$, which has same view point with $I_2$, and thus similar appearance.
We next extract a SIFT descriptor $f'_{1,i}$ on $I'_1$ for bounded search rather than using the original $f_{1,i}$.
Two example patches are provided in Fig.~\ref{fig:warped_patches}:
although $I_1$ and $I_2$ have significant appearance variation, since they are taken from different rows, the warped $I'_1$ has a very similar appearance to $I_2$, which makes  SIFT descriptor matching possible.

\begin{algorithm}
\caption{Robust Data Association}
\DontPrintSemicolon
\SetAlgoLined
\SetKwInOut{Input}{Input}\SetKwInOut{Output}{Output}
\Input{Image $I_1, I_2$, Camera $C_1, C_2$, Landmarks $L_1$}
\Output{Set of matched feature point pairs $P_{1,2}$}
set match set $P_{1,2} = \emptyset$\;
\ForEach{$l_i \in L_1$}{
    back project $l_i$ to $C_2 \rightarrow p_{2,i}$\;
    \eIf{$C_1$ and $C_2$ baseline length $<$ threshold}{
    		$f'_{1,i} = f_{1,i}$\;
    }{
    		calculate homography $\mathbf{H}_{1,2}$ use Eq.~\ref{eq:H_12}\;
    		use $\mathbf{H}_{1,2}$ warp $I_1 \rightarrow I'_1$\;
    		calculate SIFT descriptor at $p_{2,i}$ on $I'_1 \rightarrow f'_{1,i}$\;
    	}
    set $l_i$'s match set $P_{i} = \emptyset$\;
	\ForEach{feature point $f_{2,j} \in I_2$}{
		\If{$f_{2,j}$ in bounding box of $p_{2,i}$} {
			insert $[ f'_{1,i}, f_{2,j} ] \rightarrow P_{i}$\;
		}	
	}
	find min $L2$ of SIFT descriptor in $P_{i} \rightarrow [ f'_{1,i}, f_{2,k} ]$ \;
	insert $[ f_{1,i}, f_{2,k} ]$ into $P_{1,2}$\;
}
\textit{RANSAC\_8pt\_reject\_outlier}($P_{1,2}$)\;
\label{algo:data_association}
\end{algorithm}

\begin{figure}
\centering
\begin{subfigure}[b]{0.45\textwidth}
\centering
\includegraphics[width=1\linewidth]{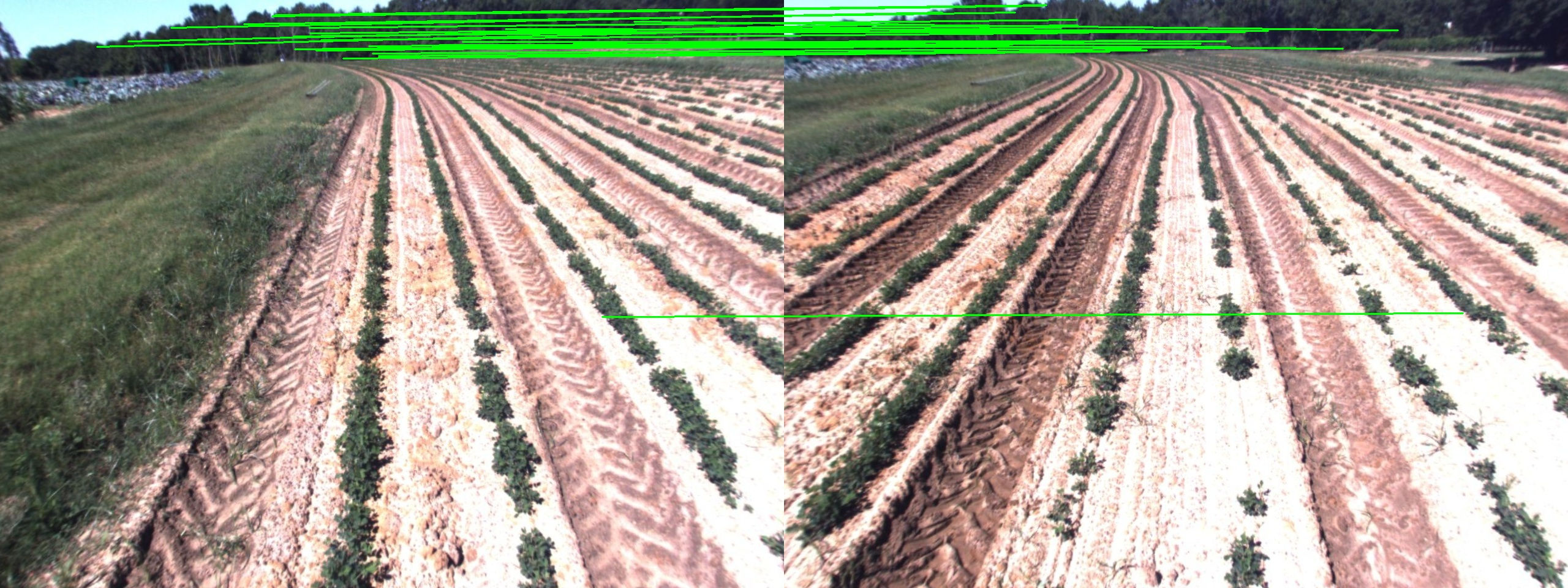}
\caption{data association by FLANN and 8-point RANSAC}
\end{subfigure}\\
\vspace*{1mm}
\begin{subfigure}[b]{0.45\textwidth}
\centering
\includegraphics[width=1\linewidth]{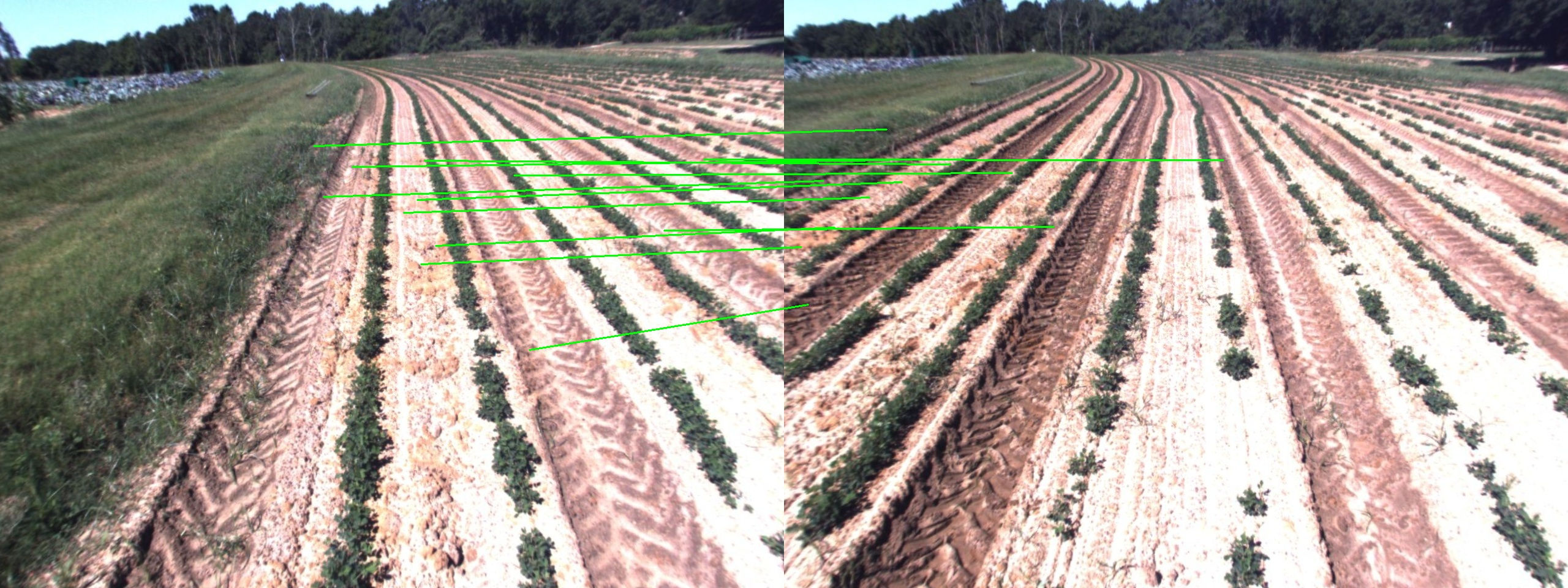}
\caption{data association by proposed approach}
\end{subfigure}
\protect\caption{
Data association results of a image pair between 1st and 3rd row of June 9.
Best viewed in digital.
\label{fig:cross_row_matches}}
\end{figure}

\begin{figure}
\centering
\includegraphics[width=0.45\textwidth]{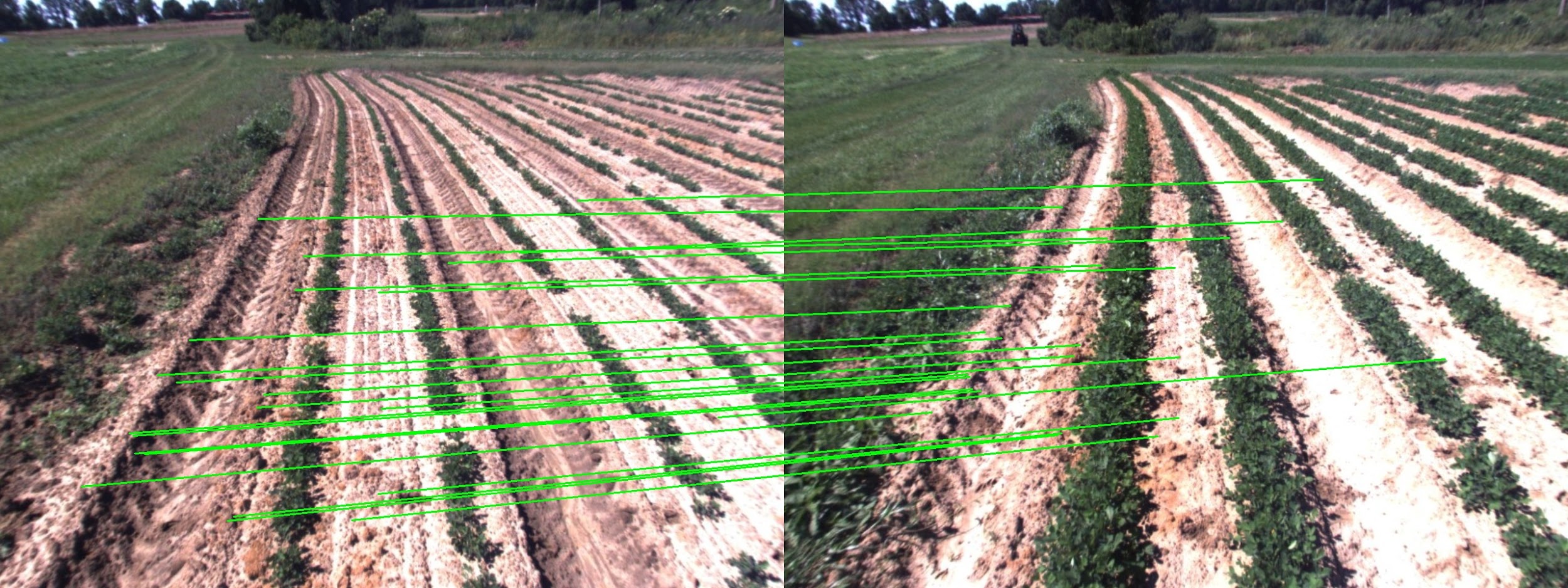}
\protect\caption{
Data association results of a image pair between 1st row of June 9 and June 20.
Best viewed in digital.
\label{fig:cross_time_matches}}
\end{figure}

The full data association pipeline is summarized in Algo.~\ref{algo:data_association}.
We estimate the normal vector $\mathbf{n}_i$ by a local landmark point cloud around $l_i$ in $L_1$.
Homography image warping is only enabled when the baseline between $C_1$ and $C_2$ is longer than a threshold, in our system this is set to $0.5$m.
After getting all nearest neighbour feature matches by back projection bounded search, a final outlier rejection is performed by 8-point RANSAC.

Experiments validate the performance and robustness of proposed approach.
A cross-row (1st vs. 3rd row) data association result is shown in Fig.~\ref{fig:cross_row_matches}.
The naive FLANN+RANSAC approach can only recover feature matches on the top, where far away objects do not change their appearance, and it fails to register any crops correctly in the field. However, the proposed approach can register feature points in the field, with significant changes of appearance. A successful cross session matching result is also shown in Fig.~\ref{fig:cross_time_matches}.

\subsection{4D Reconstruction}

\begin{figure}[!t]
\centering
{\includegraphics[width=1.0\columnwidth]{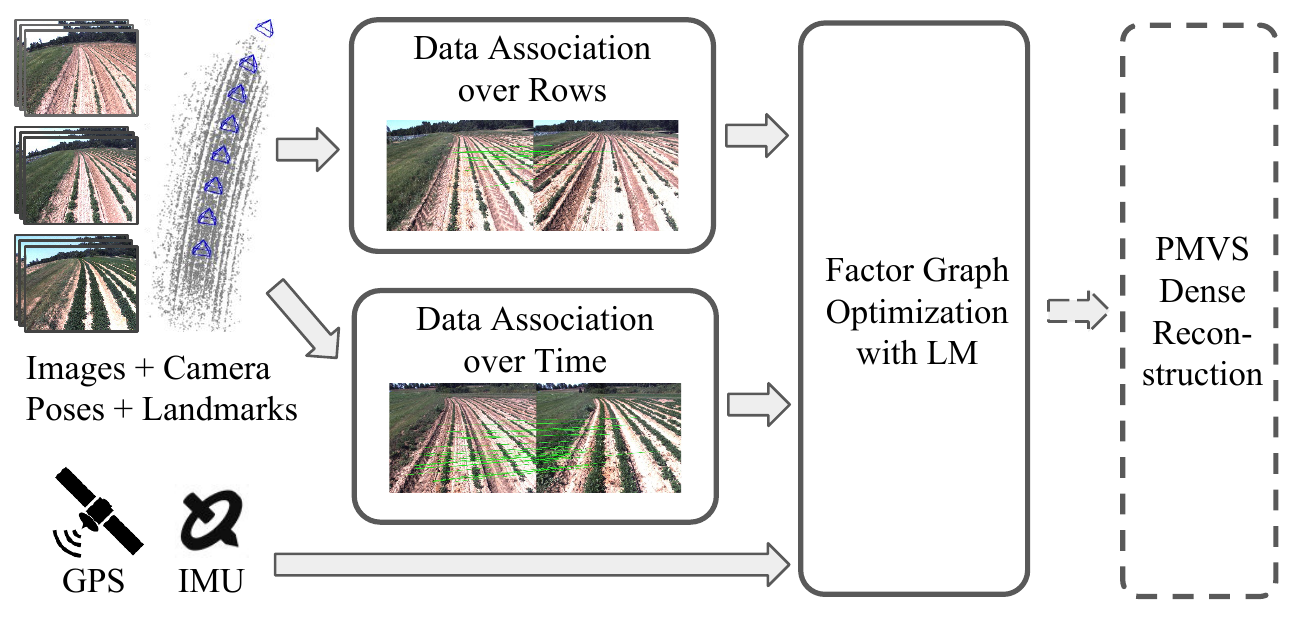}}
\caption{Overview of 4D reconstruction pipeline. Dash box of PMVS dense reconstruction step means it is optional.}
\label{fig:4d_pipeline}
\end{figure}

The third of last part of our pipeline is a 4D reconstrcution module.
The complete 4D reconstruction pipeline is illustrated  in Fig.~\ref{fig:4d_pipeline}.
We define the goal of 4D optimization as jointly estimating all camera states $X = \bigcup_{t_i \in T, r_j \in R} X^{\langle t_i, r_j \rangle}$ and all landmarks $L = \bigcup_{t_i \in T, r_j \in R} L^{\langle t_i, r_j \rangle}$, where $R$ and $T$ are set of rows and sessions, respectively.
The measurements  $Z = \bigcup_{t_i \in T, r_j \in R} Z^{\langle t_i, r_j \rangle} \cup Z_{cr}$ includes all single row information as well as data association measurements $Z_{cr}$ that connect rows across space and time.

\begin{figure}
\centering
\begin{subfigure}[b]{0.20\textwidth}
\centering
\includegraphics[width=1\linewidth]{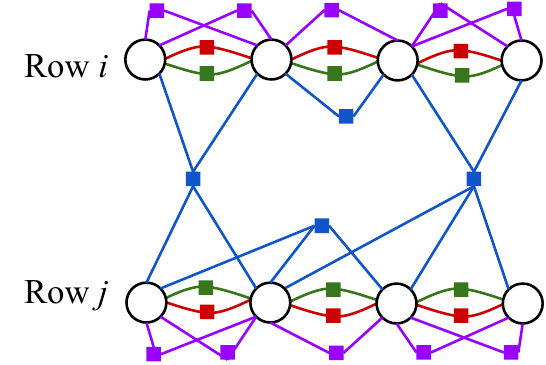}
\caption{4D factor graph}
\end{subfigure}
\hfill
\begin{subfigure}[b]{0.25\textwidth}
\centering
\includegraphics[width=1\linewidth]{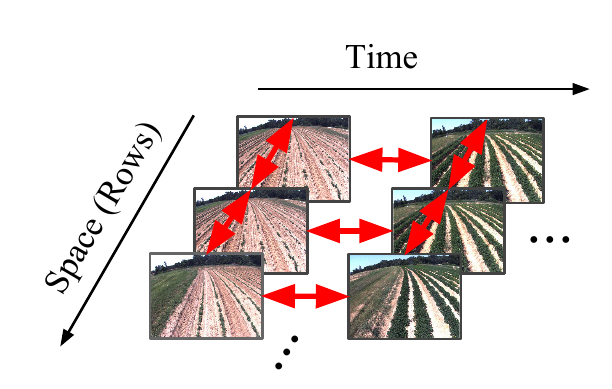}
\caption{4D data association pattern}
\end{subfigure}
\protect\caption{
(a) Factor graph of two rows with data association, connected vision factors are shared (matched) landmarks in two rows.
(b) Data association pattern of 4D reconstruction. 
\label{fig:4d_data_association}}
\end{figure}

Similar to the multi-sensor SLAM, we also formulate the joint probability of all camera states
\begin{equation}
p(X | Z) \propto \prod_{t_i \in T} \prod_{r_j \in R} \phi(X^{\langle t_i, r_j \rangle}) \prod^{H}_{h=1} \phi(X_{cr,h}),
\end{equation}
where $H$ here is the size of $Z_{cr}$, and $X_{cr,h}$ is the set of states $h$th measurement of $Z_{cr}$ involved. 
This joint probability can be expressed as a factor graph, shown in Fig.~\ref{fig:4d_data_association}(a).
The first part of the joint probability consists of the factor graphs from all single rows. And the second part is the cross-row and cross-session measurements $Z_{cr}$, which are vision factors generated from cross-row and cross-session data association.
We call the added factors \emph{shared landmarks}, since they are shared by two (or possibly more) rows, and they have two (or more) sessions associated with them.

Solving the MAP estimation problem results in estimated camera states
\begin{equation}
\hat{X} = \argmax_{X} p(X | Z).
\end{equation}
We use the Levenberg-Marquardt algorithm to solve the optimization problem, with initialization from the result of multi-sensor SLAM.
Outlier rejection of vision factors~\cite{Carlone14icra} is also enabled during optimization, to reject possible false positive feature matches from cross-row and cross-session data association.
Landmarks $\hat{L}$ are estimated by triangulation given estimated camera poses.

Data association is performed across different rows and times to get $Z_{cr}$. 
Exhaustive search between all row pairs is not necessary, since rows are not visible from each other in the images, and large timespans makes matches between images difficult to calculate.
In our approach we only match rows next to each other in either the space domain (near-by rows in the field), or the time domain (near-by date), as shown in Fig.~\ref{fig:4d_data_association}(b).

The point cloud $\hat{L}$ is relatively sparse, since it comes from a feature-base SLAM pipeline, where only points with distinct appearance are accepted as landmarks (in our system SIFT key points are accepted). 
An optional solution is to use PMVS~\cite{Furukawa10pami}, which takes estimated camera states $\hat{X}$ to reconstruct dense point clouds. 

\section{Evaluation}

\subsection{Dataset} \label{sec:dataset}

To evaluate the performance of our approach with real-world data, we collected a field dataset with large spatial and temporal scales.
Existing datasets with both large scale spatial and temporal information include the CMU dataset~\cite{CMUdataset}, the MIT dataset~\cite{Fallon2012rssws}, and the UMich dataset~\cite{Carlevaris16ijrr}. However, all of these datasets are collected in urban environments, and are not suitable for precision agriculture applications.

\begin{figure}
\centering
\begin{minipage}[b]{0.40\columnwidth}
\centering
\includegraphics[width=1\linewidth]{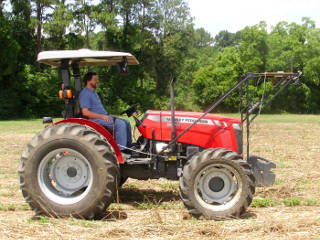}
\par\bigskip
\includegraphics[width=1\linewidth]{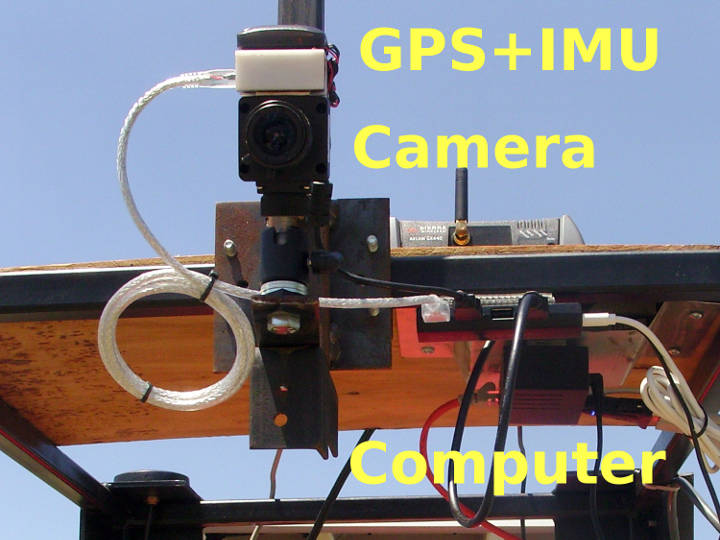}
\end{minipage}
\hfill
\begin{minipage}[b]{0.58\columnwidth}
\includegraphics[width=1\linewidth]{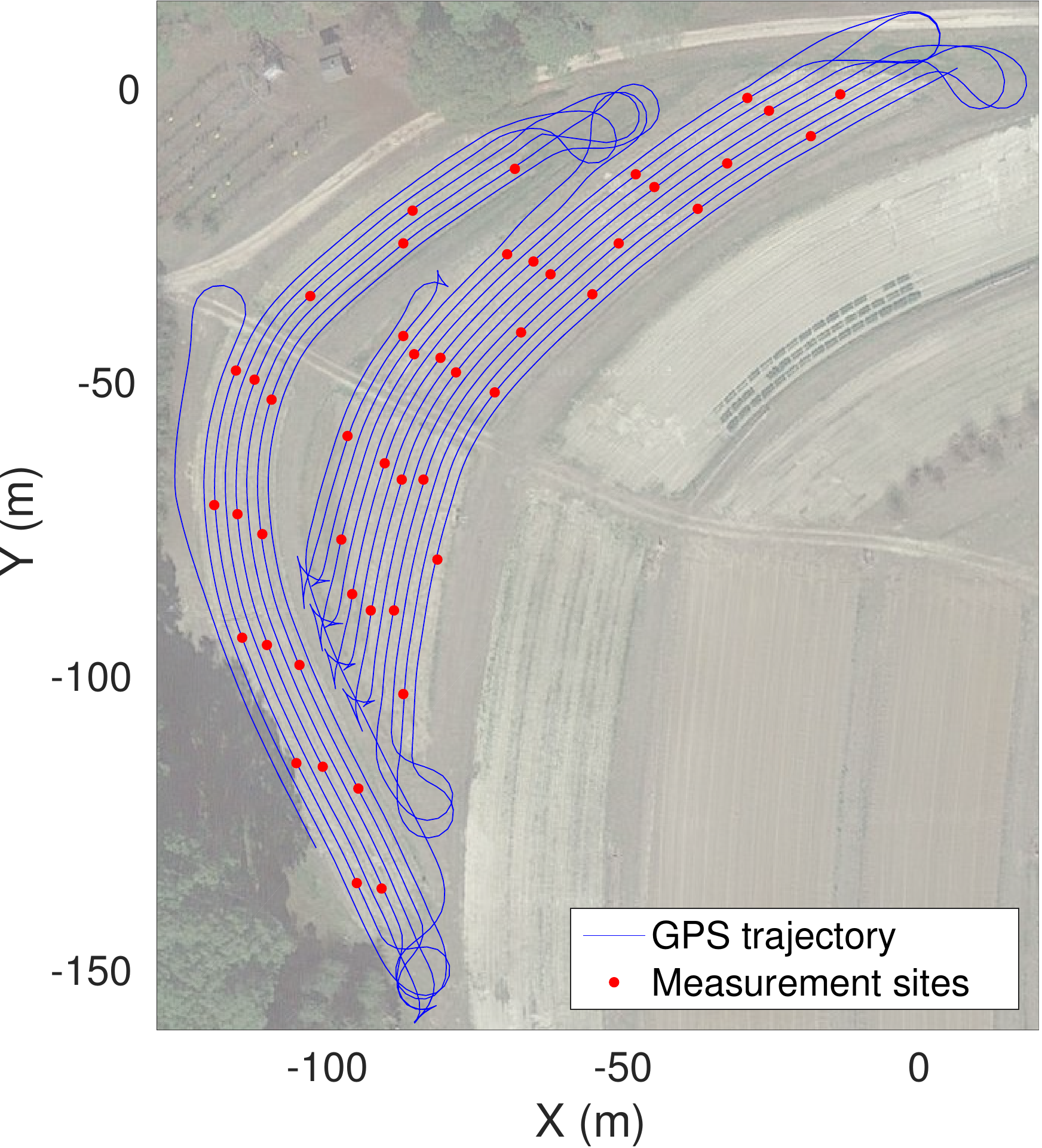}
\end{minipage}
\caption{Top left is the tractor collected the dataset; 
Down left shows sensors and computer (RTK-GPS is not shown);
Right is a sample RTK-GPS trajectory, and sites of manual measurements are taken, overlay on Google Maps.}
\label{fig:field_dataset}
\end{figure}

The dataset was collected from a field located in Tifton, GA, USA. 
The size of the field is about 150m$\times$120m, and it contains total 21 rows of peanut plants.
The map of the field is shown in Fig.~\ref{fig:field_dataset}.

We use a ground vehicle (tractor) equipped with multiple sensors, shown in Fig.~\ref{fig:field_dataset}, to collect all of the sensor data.
The equipped sensors include: (1) a Point Grey monocular global shutter camera, 1280$\times$960 color images are streamed at 7.5Hz, (2) a 9DoF IMU with compass, acceleration and angular rate are streamed at 167Hz, and magnetic field data is streamed at 110Hz, (3) a high accuracy RTK-GPS, and a low accuracy GPS, both of them stream latitude and longitude data at 5Hz.
No hardware synchronization is used.
All data are stored in a SSD by an on-board computer.

\begin{figure}
\centering
\includegraphics[width=0.24\linewidth]{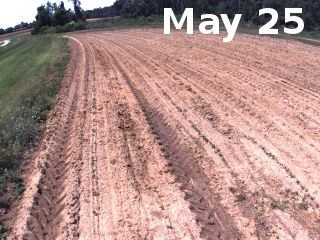}
\includegraphics[width=0.24\linewidth]{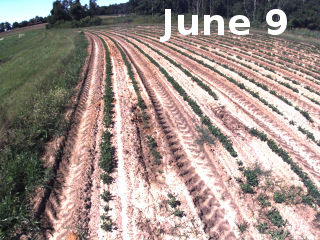}
\includegraphics[width=0.24\linewidth]{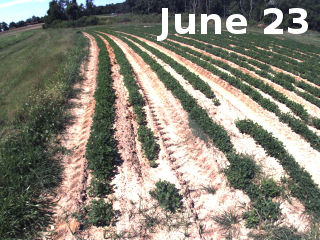}
\includegraphics[width=0.24\linewidth]{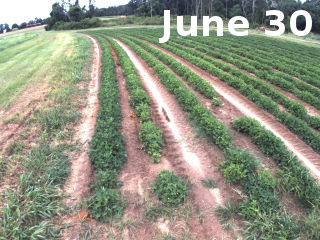}\\
\vspace*{1mm}
\includegraphics[width=0.24\linewidth]{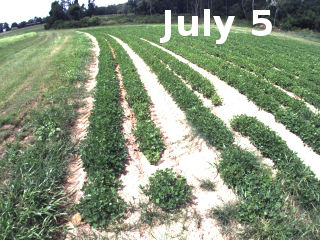}
\includegraphics[width=0.24\linewidth]{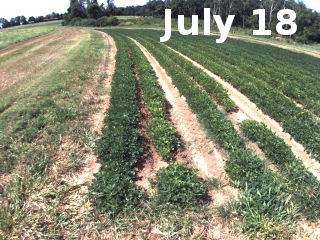}
\includegraphics[width=0.24\linewidth]{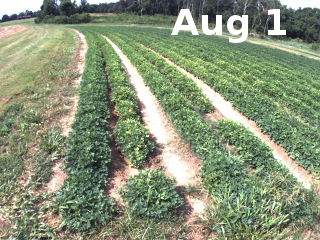}
\includegraphics[width=0.24\linewidth]{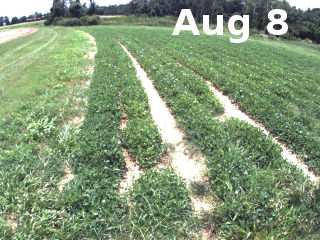}
\protect\caption{
Eight sample images taken at approximately same location in the field, dates taken are marked on images.
\label{fig:field_dataset_imgs}}
\end{figure}

We recorded a complete season of peanut growth which started May 25, 2016 and completed Aug 22, 2016, right before harvest.
The data collection had a total of 23 sessions over 89 days, approximately two per week, with a few exceptions due to severe weather.
Example images of different dates are shown in Fig.~\ref{fig:field_dataset_imgs}.
Each session lasted about 40 minutes, and consisted of the tractor driving about 3.8km in the field.

In addition to sensor data, ground truth crop properties (height and leaf chlorophyll) at multiple sampling sites in the field were measured weekly by a human operator. 
There were a total of 47 measuring sites, as shown in Fig.~\ref{fig:field_dataset}.

\subsection{Results}

We ran the proposed 4D reconstruction approach on the peanut field dataset.
We implemented the proposed approach with the GTSAM C++ library.\footnote{\url{https://bitbucket.org/gtborg/gtsam}}
We used RTK-GPS data from the dataset as GPS input, and ignored lower accuracy GPS data.
Since the peanut field contains two sub-fields with little overlap (see Fig.~\ref{fig:field_dataset}), the two sub-fields were reconstructed independently and aligned by GPS. 
Since the tractor runs back and forth in the field, we only use rows in which the tractor driving south (odd rows), to avoid misalignment with reconstruction results from even rows.
Example densely reconstructed 4D results are shown in Fig.~\ref{fig:4d_results_1}.

Although Fig.~\ref{fig:4d_results_1} shows that the 3D reconstruction results for each single session qualitatively appear accurate, to make these results useful to precision agriculture applications, are interested in evaluating the approach quantitatively. In particular we wanted to answer the following questions:
\begin{itemize}
\item Are these 3D results correctly aligned in space?
\item Are these 3D results useful for measuring geometric properties of plants useful for crop monitoring (height, width, etc.) ?
\end{itemize}

\begin{figure}[!t]
\centering
{\includegraphics[width=1.0\columnwidth]{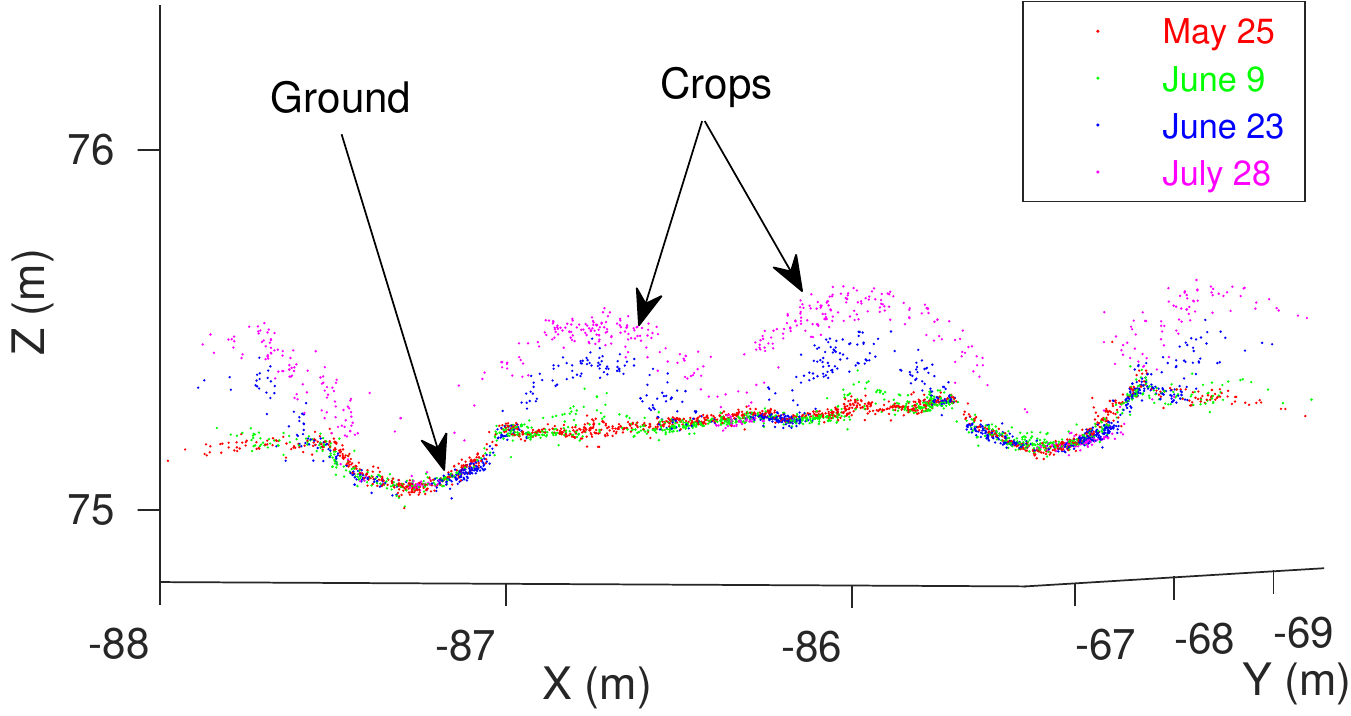}}
\caption{Cross section of part of the sparse 4D reconstruction results at 3rd row. Only 4 sessions are shown to keep figure clear. Best viewed in digital.}
\label{fig:4d_cross_section}
\end{figure}

To answer the first question, we visualize the 4D model by showing all 3D point clouds together. 
We visualize part of the 4D sparse reconstruction result in Fig.~\ref{fig:4d_cross_section}. Point clouds from different dates are marked in different colors. 
We can see from the cross section that the ground surface point clouds from different sessions are well aligned, which shows that all of the 3D point clouds from different dates are well registered into a single coordinate frame. This suggests that we are building a true 4D result. 
We can see the growth of the peanut plants, as the point cloud shows `Matryoshka doll' like structure, earlier crop point clouds are contained with in point clouds of later sessions.

To answer the second question, we show some preliminary crop analysis results using  reconstructed 4D point clouds, compared with ground truth manual measurements.
We setup a simple pipeline to estimate height of peanut plants from sparse reconstructed 4D point clouds at multiple sites, by first estimating the local ground plane by RANSAC from May 25's point cloud (when peanuts are small and ground plane is well reconstructed), second separate peanut's point clouds by color (using RGB values), and finally estimate the distance from peanut canopy's top to ground plane.

\begin{figure*}
\centering
\includegraphics[width=0.32\linewidth]{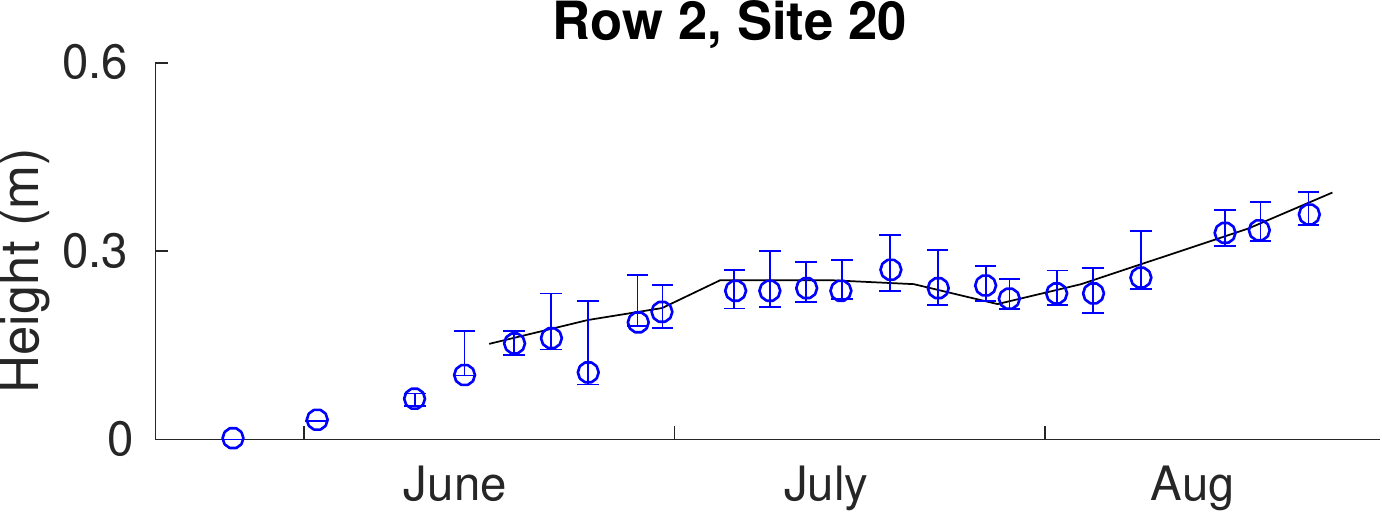}
\includegraphics[width=0.32\linewidth]{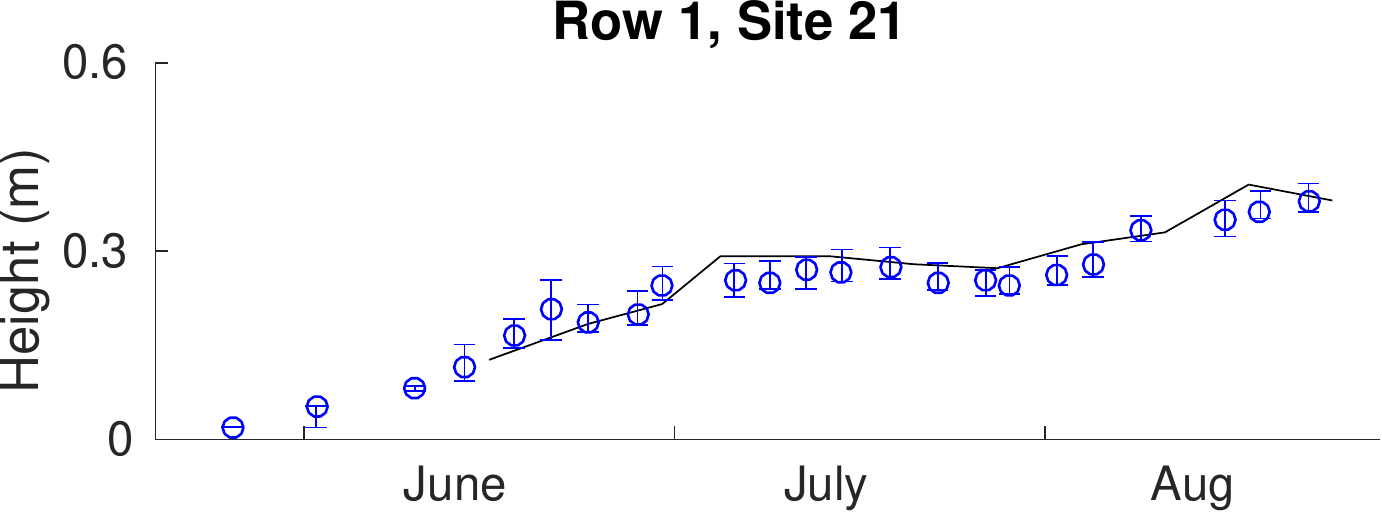}
\includegraphics[width=0.32\linewidth]{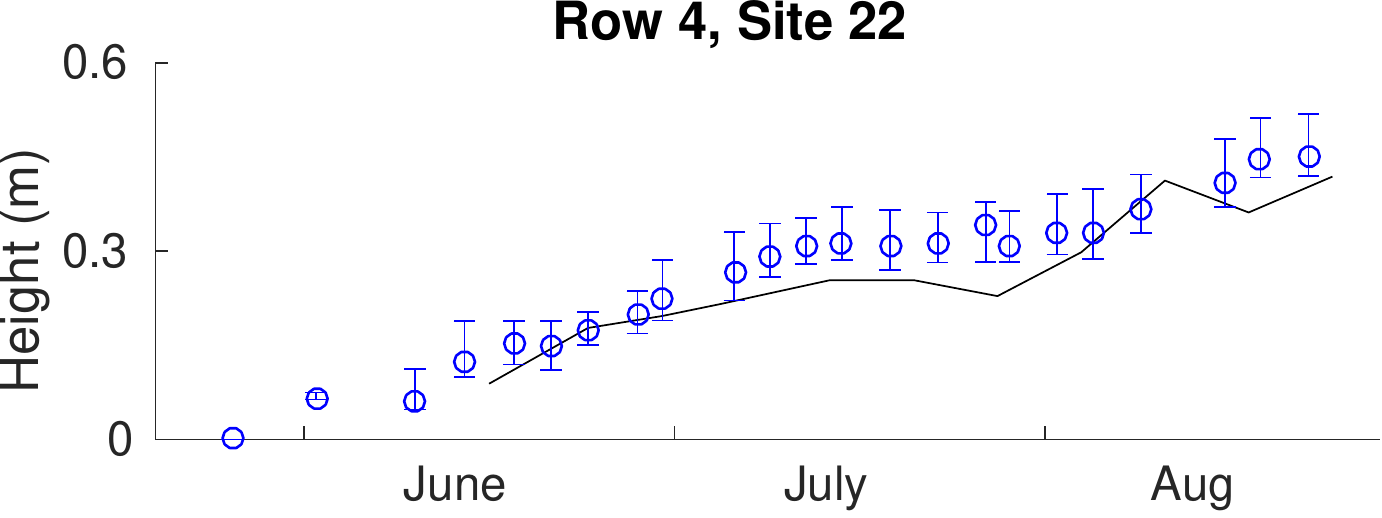}\\
\vspace*{1mm}
\includegraphics[width=0.32\linewidth]{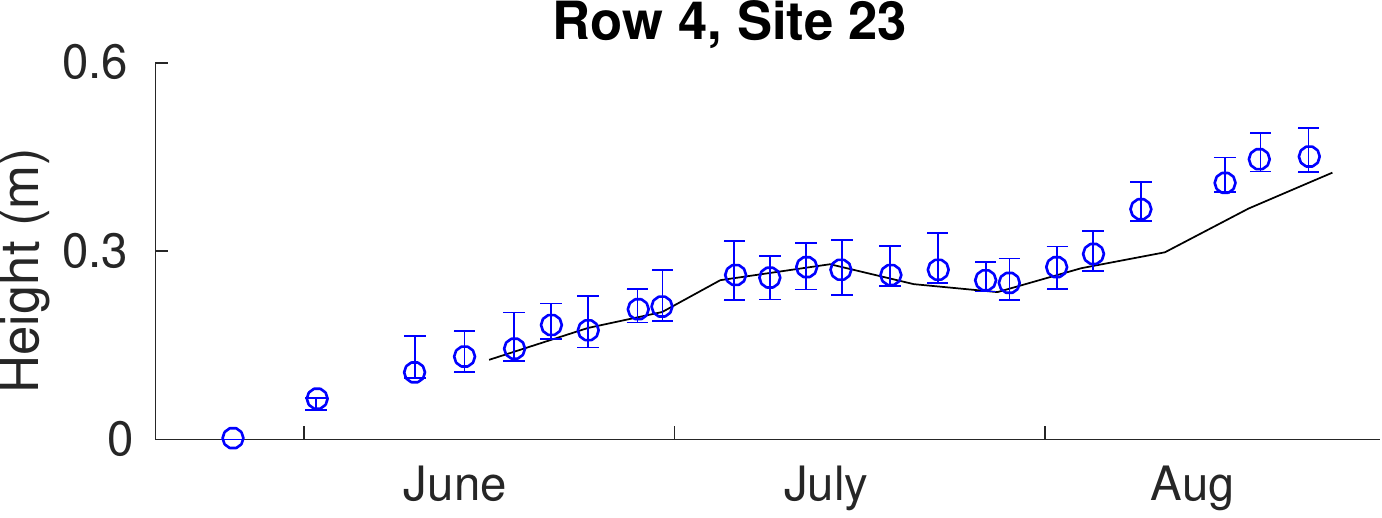}
\includegraphics[width=0.32\linewidth]{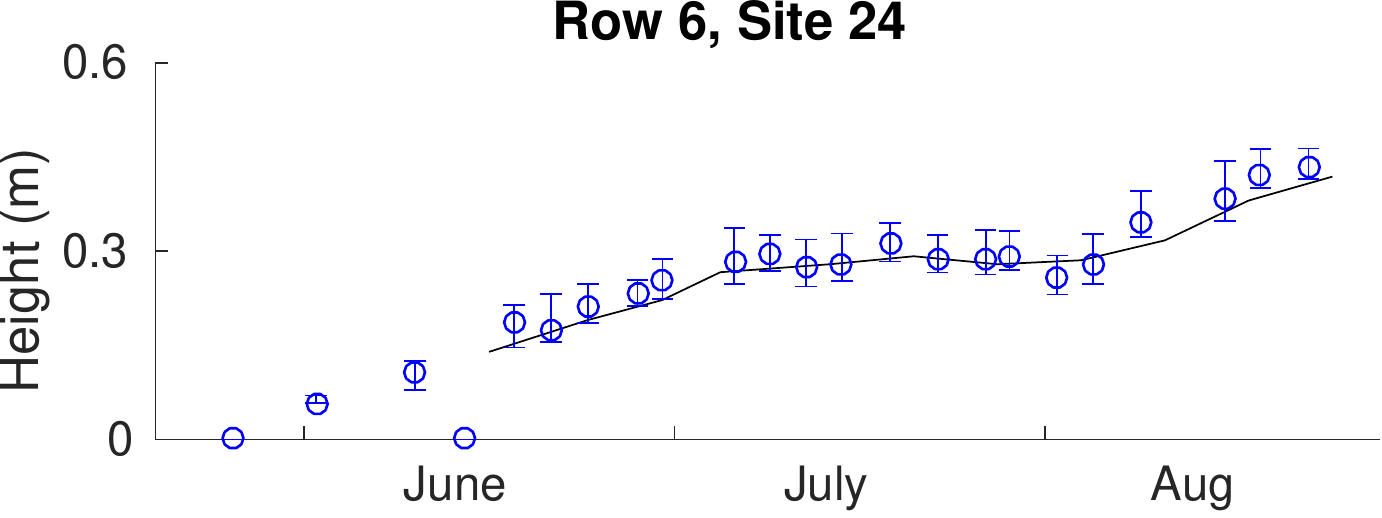}
\includegraphics[width=0.32\linewidth]{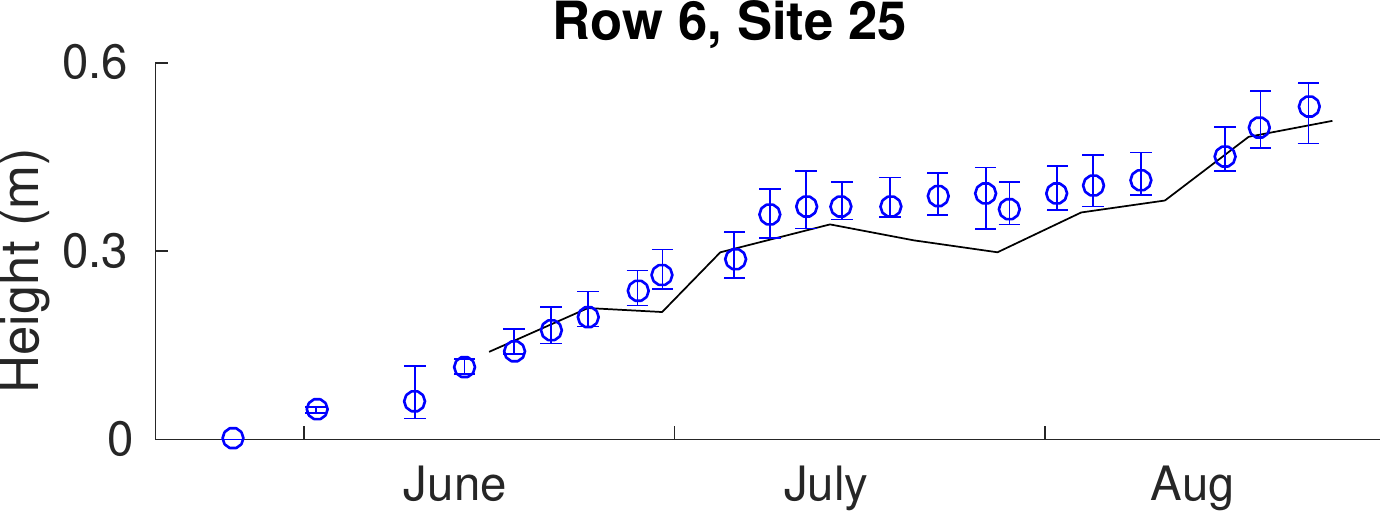}\\
\vspace*{1mm}
\includegraphics[width=0.32\linewidth]{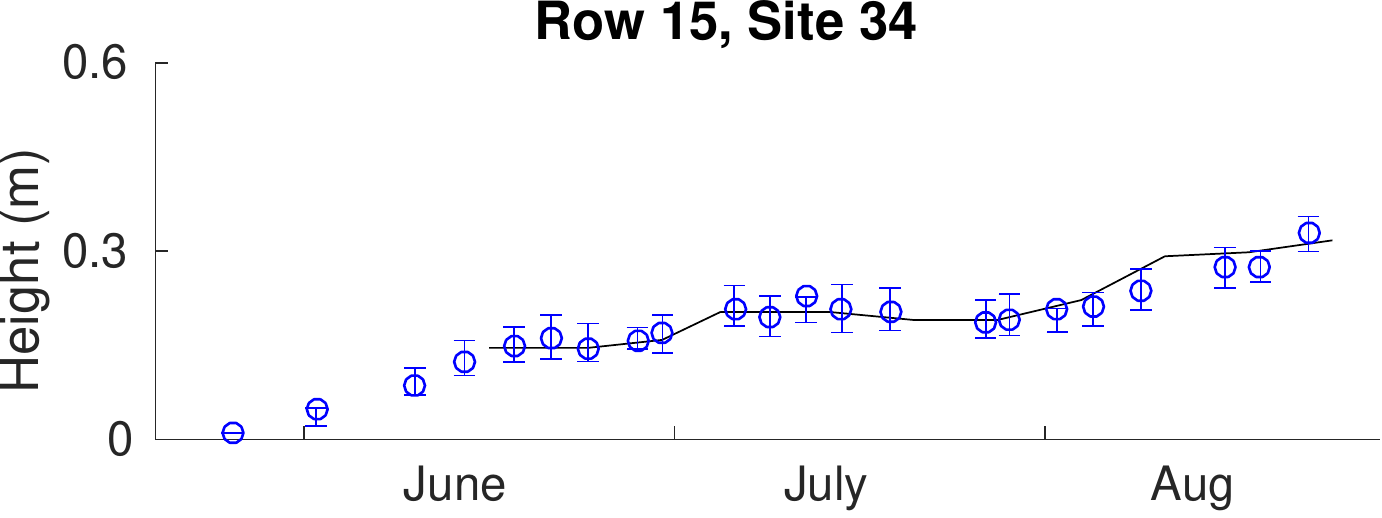}
\includegraphics[width=0.32\linewidth]{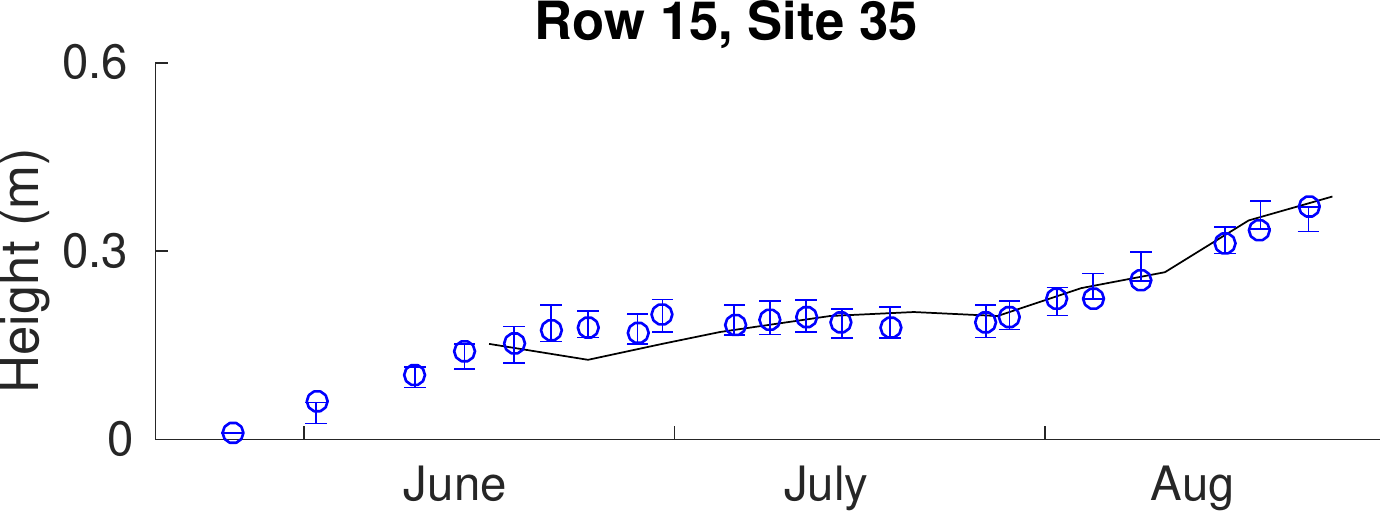}
\includegraphics[width=0.32\linewidth]{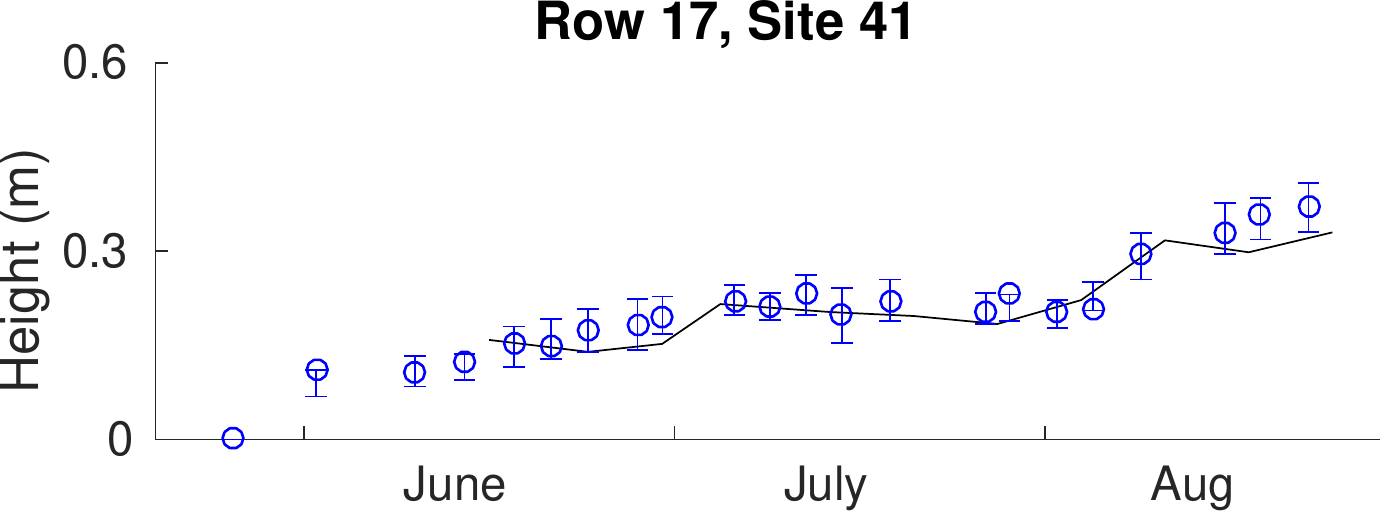}\\
\vspace*{1mm}
\includegraphics[width=0.32\linewidth]{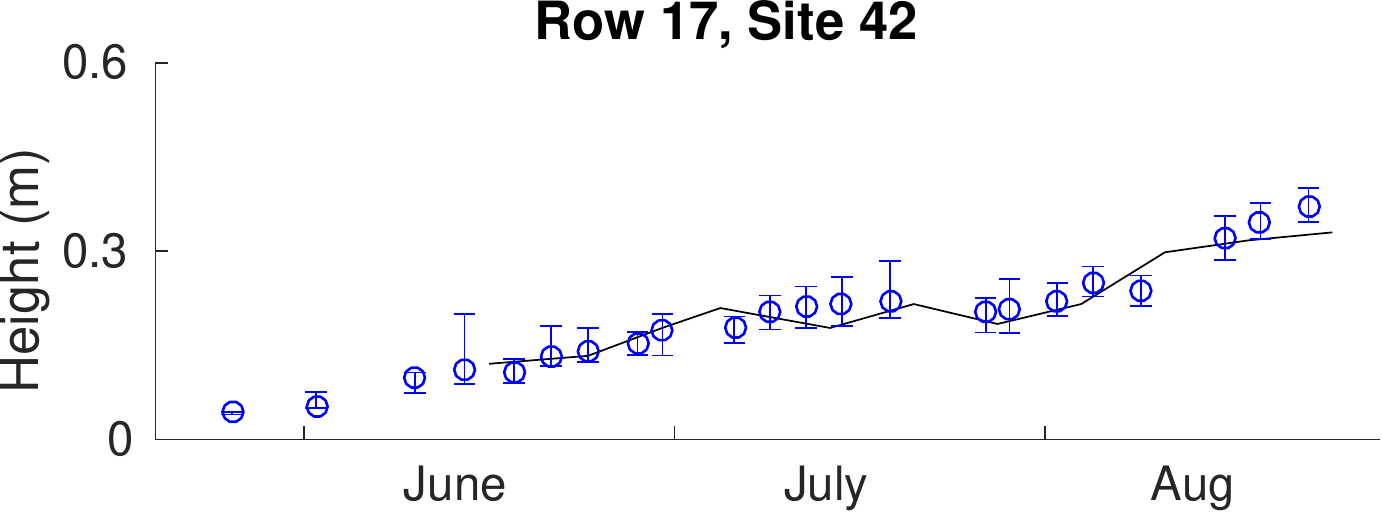}
\includegraphics[width=0.32\linewidth]{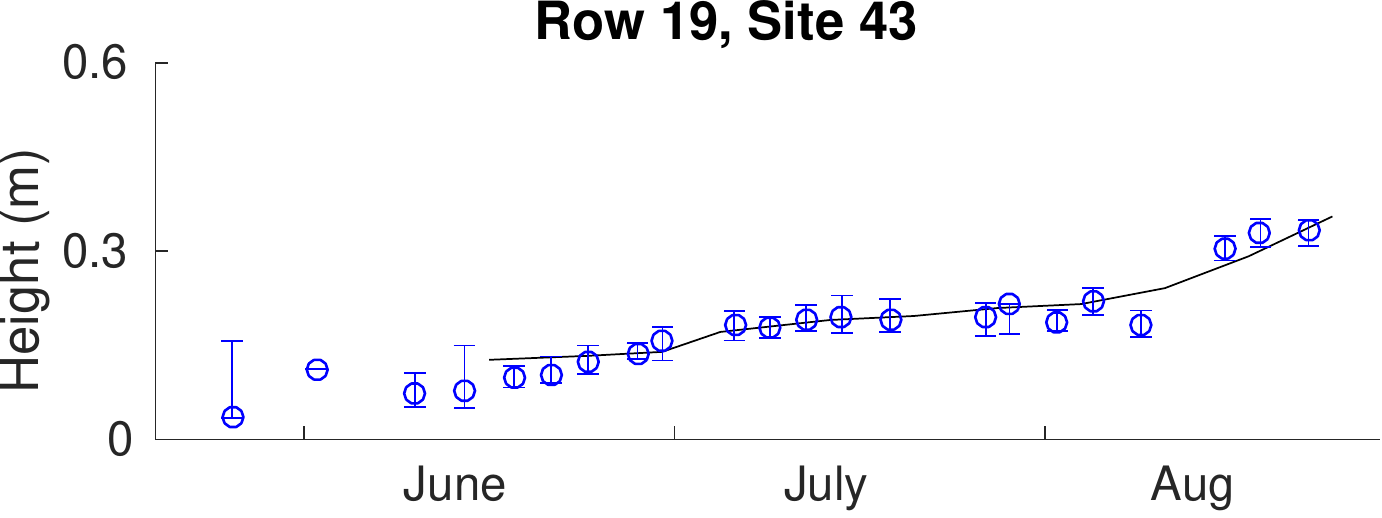}
\includegraphics[width=0.32\linewidth]{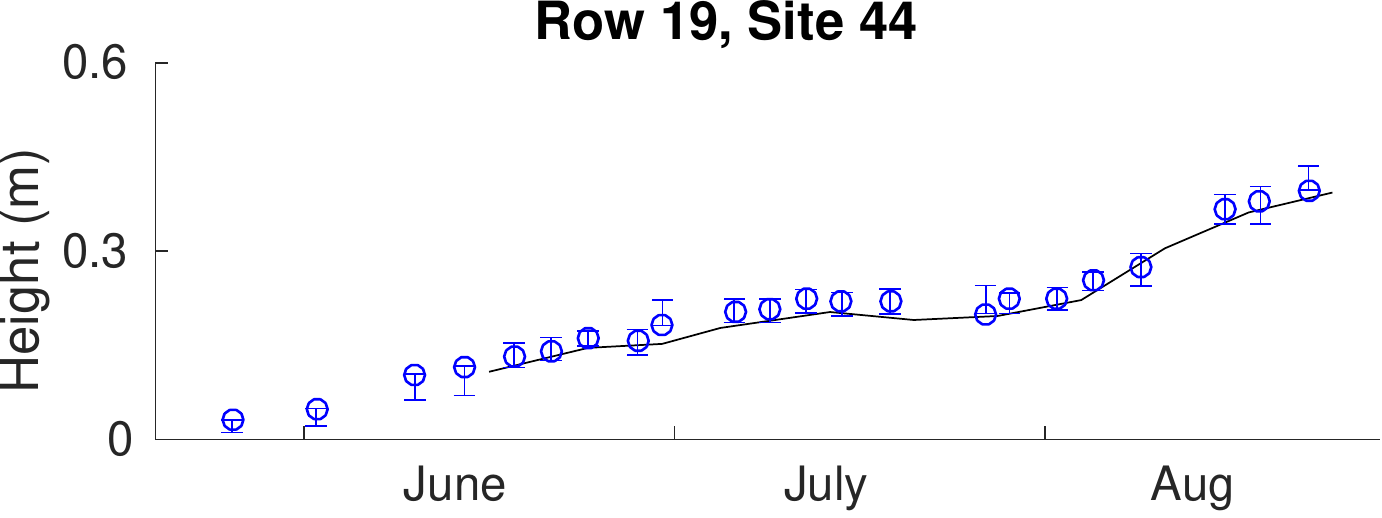}
\protect\caption{
Estimated peanut heights at 12 sampling sites in blue, with ground truth manual measurements in black lines.\label{fig:field_height_results}}
\end{figure*}

Preliminary height estimations of twelve sampling sites are shown in Fig.~\ref{fig:field_height_results}.
With the exception of sites 22 and 25, which have slightly biased estimated heights due to poor RANSAC ground plane estimations, results of the sites meet the ground truth measurements well. 
This shows that we can compute reasonable height estimates even with a simple method, and proves that the 4D reconstruction results contain correct geometric statistics.

\section{Conclusion and Future Work}

In this paper we address the 4D reconstruction problem for crop monitoring applications. 
The outcome of the proposed 4D approach is a set of 3D point clouds, with pleasing visual appearance and correct geometric properties.
A robust data association algorithm is also developed to address the problems inherent in matching image features from difference dates and different view points, while performing 4D reconstruction.
A side product of this paper is a high quality field dataset for testing crop monitoring applications, which will be released to the public.

Although we show some preliminary results of crop height analysis, this paper is mainly solving the problem of reconstructing the 4D point clouds, not the point cloud analysis.
We leave the use of existing methods  for analyzing point clouds~\cite{Li13analyzing,Carlone15icraws} or proposing more sophisticated point cloud analysis, as future work.

One possible extension to the proposed approach is further improvement of the data association process.
Although our data association method outperforms existing approaches, it still needs assistance from high accuracy RTK-GPS as prior information, and we experience a few failure cases due to thunderstorms that wash out most features on the ground between sessions. 

Although the system is designed for precision agriculture applications, any tasks satisfying the assumptions mentioned in Sec.~\ref{sec:method} can utilize the proposed system.
For example, we can use this system to build maps for autonomous driving, if we are not aiming to reconstruct dynamic objects, since every street can be treated as a single `row'.

\section*{Acknowledgement}

This work is supported by National Institute of Food and Agriculture, U.S. Department of Agriculture, under award number 2014-67021-22556.

\bibliographystyle{ieeetr}
\bibliography{refs}

\begin{thebibliography}{10}

\bibitem{Rembold13using}
F.~Rembold, C.~Atzberger, I.~Savin, and O.~Rojas, ``Using low resolution
  satellite imagery for yield prediction and yield anomaly detection,'' {\em
  Remote Sensing}, vol.~5, no.~4, pp.~1704--1733, 2013.

\bibitem{Bryson10airborne}
M.~Bryson, A.~Reid, F.~Ramos, and S.~Sukkarieh, ``Airborne vision-based mapping
  and classification of large farmland environments,'' {\em J. of Field
  Robotics}, vol.~27, no.~5, pp.~632--655, 2010.

\bibitem{Anthony14iros}
D.~Anthony, S.~Elbaum, A.~Lorenz, and C.~Detweiler, ``On crop height estimation
  with {UAVs},'' in {\em IEEE/RSJ Intl. Conf. on Intelligent Robots and Systems
  (IROS)}, 2014.

\bibitem{Das15devices}
J.~Das, G.~Cross, C.~Qu, A.~Makineni, P.~Tokekar, Y.~Mulgaonkar, and V.~Kumar,
  ``Devices, systems, and methods for automated monitoring enabling precision
  agriculture,'' in {\em IEEE Intl. Conf. on Automation Science and Engineering
  (CASE)}, 2015.

\bibitem{Zainuddin16cspa}
K.~Zainuddin, M.~Jaffri, M.~Zainal, N.~Ghazali, and A.~Samad, ``Verification
  test on ability to use low-cost {UAV} for quantifying tree height,'' in {\em
  IEEE Intl. Colloquium on Signal Processing \& Its Applications (CSPA)}, 2016.

\bibitem{Sarkar16towards}
S.~K. Sarkar, J.~Das, R.~Ehsani, and V.~Kumar, ``Towards autonomous
  phytopathology: Outcomes and challenges of citrus greening disease detection
  through close-range remote sensing,'' in {\em IEEE Intl. Conf. on Robotics
  and Automation (ICRA)}, 2016.

\bibitem{Hague06automated}
T.~Hague, N.~Tillett, and H.~Wheeler, ``Automated crop and weed monitoring in
  widely spaced cereals,'' {\em Precision Agriculture}, vol.~7, no.~1,
  pp.~21--32, 2006.

\bibitem{Lalonde06automatic}
J.-F. Lalonde, N.~Vandapel, and M.~Hebert, ``Automatic three-dimensional point
  cloud processing for forest inventory,'' {\em Robotics Institute}, p.~334,
  2006.

\bibitem{Singh10comprehensive}
S.~Singh, M.~Bergerman, J.~Cannons, B.~Grocholsky, B.~Hamner, G.~Holguin,
  L.~Hull, V.~Jones, G.~Kantor, H.~Koselka, {\em et~al.}, ``Comprehensive
  automation for specialty crops: Year 1 results and lessons learned,'' {\em
  Intelligent Service Robotics}, vol.~3, no.~4, pp.~245--262, 2010.

\bibitem{Nuske14automated}
S.~Nuske, K.~Wilshusen, S.~Achar, L.~Yoder, S.~Narasimhan, and S.~Singh,
  ``Automated visual yield estimation in vineyards,'' {\em J. of Field
  Robotics}, vol.~31, no.~5, pp.~837--860, 2014.

\bibitem{Font15vineyard}
D.~Font, M.~Tresanchez, D.~Mart{\'\i}nez, J.~Moreno, E.~Clotet, and
  J.~Palac{\'\i}n, ``Vineyard yield estimation based on the analysis of high
  resolution images obtained with artificial illumination at night,'' {\em
  Sensors}, vol.~15, no.~4, pp.~8284--8301, 2015.

\bibitem{Sa16deepfruits}
I.~Sa, Z.~Ge, F.~Dayoub, B.~Upcroft, T.~Perez, and C.~McCool, ``Deepfruits: A
  fruit detection system using deep neural networks,'' {\em Sensors}, vol.~16,
  no.~8, p.~1222, 2016.

\bibitem{Agarwal09iccv}
S.~Agarwal, N.~Snavely, I.~Simon, S.~M. Seitz, and R.~Szeliski, ``Building
  {R}ome in a day,'' in {\em Intl. Conf. on Computer Vision (ICCV)}, 2009.

\bibitem{Furukawa10pami}
Y.~Furukawa and J.~Ponce, ``Accurate, dense, and robust multi-view
  stereopsis,'' {\em {IEEE} Trans. Pattern Anal. Machine Intell.}, vol.~32,
  no.~8, pp.~1362--1376, 2010.

\bibitem{Sakurada13cvpr}
K.~Sakurada, T.~Okatani, and K.~Deguchi, ``Detecting changes in 3{D} structure
  of a scene from multi-view images captured by a vehicle-mounted camera,'' in
  {\em IEEE Conf. on Computer Vision and Pattern Recognition (CVPR)}, 2013.

\bibitem{Alcantarilla16rss}
P.~F. Alcantarilla, S.~Stent, G.~Ros, R.~Arroyo, and R.~Gherardi, ``Street-view
  change detection with deconvolutional networks,'' in {\em Robotics: Science
  and Systems (RSS)}, 2016.

\bibitem{Xiao15isprs}
W.~Xiao, B.~Vallet, M.~Br{\'e}dif, and N.~Paparoditis, ``Street environment
  change detection from mobile laser scanning point clouds,'' {\em ISPRS
  Journal of Photogrammetry and Remote Sensing}, vol.~107, pp.~38--49, 2015.

\bibitem{Pollard07cvpr}
T.~Pollard and J.~L. Mundy, ``Change detection in a 3-{D} world,'' in {\em IEEE
  Conf. on Computer Vision and Pattern Recognition (CVPR)}, 2007.

\bibitem{Ulusoy14eccv}
A.~O. Ulusoy and J.~L. Mundy, ``Image-based 4-{D} reconstruction using 3-{D}
  change detection,'' in {\em European Conf. on Computer Vision (ECCV)}, 2014.

\bibitem{Taneja11iccv}
A.~Taneja, L.~Ballan, and M.~Pollefeys, ``Image based detection of geometric
  changes in urban environments,'' in {\em Intl. Conf. on Computer Vision
  (ICCV)}, 2011.

\bibitem{Taneja13cvpr}
A.~Taneja, L.~Ballan, and M.~Pollefeys, ``City-scale change detection in
  cadastral 3{D} models using images,'' in {\em IEEE Conf. on Computer Vision
  and Pattern Recognition (CVPR)}, 2013.

\bibitem{Taneja15pami}
A.~Taneja, L.~Ballan, and M.~Pollefeys, ``Geometric change detection in urban
  environments using images,'' {\em {IEEE} Trans. Pattern Anal. Machine
  Intell.}, vol.~37, no.~11, pp.~2193--2206, 2015.

\bibitem{Golparvar11iccvws}
M.~Golparvar-Fard, F.~Pena-Mora, and S.~Savarese, ``Monitoring changes of 3{D}
  building elements from unordered photo collections,'' in {\em ICCV Workshop
  on Computer Vision for Remote Sensing of the Environment}, 2011.

\bibitem{Griffith15rssws}
S.~Griffith, F.~Dellaert, and C.~Pradalier, ``Robot-enabled lakeshore
  monitoring using visual {SLAM} and {SIFT} flow,'' in {\em RSS Workshop on
  Multi-View Geometry in Robotics}, Citeseer, 2015.

\bibitem{Griffith16bmvc}
S.~Griffith and C.~Pradalier, ``Reprojection flow for image registration across
  seasons,'' in {\em British Machine Vision Conf. (BMVC)}, 2016.

\bibitem{Milford12icranew}
M.~J. Milford and G.~F. Wyeth, ``{SeqSLAM}: Visual route-based navigation for
  sunny summer days and stormy winter nights,'' in {\em IEEE Intl. Conf. on
  Robotics and Automation (ICRA)}, 2012.

\bibitem{Naseer15ecmp}
T.~Naseer, B.~Suger, M.~Ruhnke, and W.~Burgard, ``Vision-based markov
  localization across large perceptual changes,'' in {\em European Conf. on
  Mobile Robots (ECMR)}, 2015.

\bibitem{Lowry16tro}
S.~Lowry, N.~S{\"u}nderhauf, P.~Newman, J.~J. Leonard, D.~Cox, P.~Corke, and
  M.~J. Milford, ``Visual place recognition: A survey,'' {\em {IEEE} Trans.
  Robotics}, vol.~32, no.~1, pp.~1--19, 2016.

\bibitem{Beall14ppniv}
C.~Beall and F.~Dellaert, ``Appearance-based localization across seasons in a
  metric map,'' {\em IROS Workshop on Planning, Perception and Navigation for
  Intelligent Vehicles (PPNIV)}, 2014.

\bibitem{Martin15siggraph}
R.~Martin-Brualla, D.~Gallup, and S.~M. Seitz, ``Time-lapse mining from
  internet photos,'' in {\em ACM SIGGRAPH}, 2015.

\bibitem{Martin15iccv}
R.~Martin-Brualla, D.~Gallup, and S.~M. Seitz, ``3{D} time-lapse reconstruction
  from internet photos,'' in {\em Intl. Conf. on Computer Vision (ICCV)}, 2015.

\bibitem{Schindler07cvpr}
G.~Schindler, F.~Dellaert, and S.~Kang, ``Inferring temporal order of images
  from {3D} structure,'' in {\em IEEE Conf. on Computer Vision and Pattern
  Recognition (CVPR)}, 2007.

\bibitem{Schindler10cvpr}
G.~Schindler and F.~Dellaert, ``Probabilistic temporal inference on
  reconstructed 3d scenes,'' in {\em IEEE Conf. on Computer Vision and Pattern
  Recognition (CVPR)}, 2010.

\bibitem{Matzen14eccv}
K.~Matzen and N.~Snavely, ``Scene chronology,'' in {\em European Conf. on
  Computer Vision (ECCV)}, 2014.

\bibitem{Ulusoy13iccv}
A.~O. Ulusoy, O.~Biris, and J.~L. Mundy, ``Dynamic probabilistic volumetric
  models,'' in {\em Intl. Conf. on Computer Vision (ICCV)}, 2013.

\bibitem{Lowe04ijcv}
D.~Lowe, ``Distinctive image features from scale-invariant keypoints,'' {\em
  Intl. J. of Computer Vision}, vol.~60, no.~2, pp.~91--110, 2004.

\bibitem{Muja14pami}
M.~Muja and D.~G. Lowe, ``Scalable nearest neighbor algorithms for high
  dimensional data,'' {\em {IEEE} Trans. Pattern Anal. Machine Intell.},
  vol.~36, no.~11, pp.~2227--2240, 2014.

\bibitem{Hartley04book}
R.~I. Hartley and A.~Zisserman, {\em Multiple View Geometry in Computer
  Vision}.
\newblock Cambridge University Press, second~ed., 2004.

\bibitem{Dellaert06ijrr}
F.~Dellaert and M.~Kaess, ``Square {Root} {SAM}: Simultaneous localization and
  mapping via square root information smoothing,'' {\em Intl. J. of Robotics
  Research}, vol.~25, pp.~1181--1203, Dec 2006.

\bibitem{Carlone14icra}
L.~Carlone, Z.~Kira, C.~Beall, V.~Indelman, and F.~Dellaert, ``Eliminating
  conditionally independent sets in factor graphs: A unifying perspective based
  on smart factors,'' in {\em IEEE Intl. Conf. on Robotics and Automation
  (ICRA)}, 2014.

\bibitem{Forster15rss}
C.~Forster, L.~Carlone, F.~Dellaert, and D.~Scaramuzza, ``{IMU} preintegration
  on manifold for efficient visual-inertial maximum-a-posteriori estimation,''
  {\em Robotics: Science and Systems (RSS)}, 2015.

\bibitem{Anderson15iros}
S.~Anderson and T.~D. Barfoot, ``Full {STEAM} ahead: Exactly sparse gaussian
  process regression for batch continuous-time trajectory estimation on {SE}
  (3),'' {\em IEEE/RSJ Intl. Conf. on Intelligent Robots and Systems (IROS)},
  2015.

\bibitem{Barfoot14rss}
T.~Barfoot, C.~H. Tong, and S.~Sarkka, ``Batch continuous-time trajectory
  estimation as exactly sparse gaussian process regression,'' {\em Robotics:
  Science and Systems (RSS)}, 2014.

\bibitem{Yan15isrr}
X.~Yan, V.~Indelman, and B.~Boots, ``Incremental sparse {GP} regression for
  continuous-time trajectory estimation \& mapping,'' {\em Proc. of the Intl.
  Symp. of Robotics Research (ISRR)}, 2015.

\bibitem{Kaess12ijrr}
M.~Kaess, H.~Johannsson, R.~Roberts, V.~Ila, J.~Leonard, and F.~Dellaert,
  ``{iSAM2}: Incremental smoothing and mapping using the {B}ayes tree,'' {\em
  Intl. J. of Robotics Research}, vol.~31, pp.~217--236, Feb 2012.

\bibitem{Indelman16csm}
V.~Indelman, E.~Nelson, J.~Dong, N.~Michael, and F.~Dellaert, ``Incremental
  distributed inference from arbitrary poses and unknown data association:
  Using collaborating robots to establish a common reference,'' {\em IEEE
  Control Systems}, vol.~36, no.~2, pp.~41--74, 2016.

\bibitem{CMUdataset}
H.~Badino, D.~Huber, and T.~Kanade, ``The {CMU} visual localization data set.''
  \url{http://3dvis.ri.cmu.edu/data-sets/localization}, 2011.

\bibitem{Fallon2012rssws}
M.~F. Fallon, H.~Johannsson, M.~Kaess, D.~M. Rosen, E.~Muggler, and J.~J.
  Leonard, ``Mapping the {MIT} stata center: Large-scale integrated visual and
  {RGB-D} {SLAM},'' in {\em RSS Workshop on RGB-D: Advanced Reasoning with
  Depth Cameras}, 2012.

\bibitem{Carlevaris16ijrr}
N.~Carlevaris-Bianco, A.~K. Ushani, and R.~M. Eustice, ``University of
  {Michigan} north campus long-term vision and lidar dataset,'' {\em Intl. J.
  of Robotics Research}, vol.~35, no.~9, pp.~1023--1035, 2016.

\bibitem{Li13analyzing}
Y.~Li, X.~Fan, N.~J. Mitra, D.~Chamovitz, D.~Cohen-Or, and B.~Chen, ``Analyzing
  growing plants from {4D} point cloud data,'' {\em ACM Transactions on
  Graphics (TOG) - Proceedings of ACM SIGGRAPH Asia}, vol.~32, no.~6, p.~157,
  2013.

\bibitem{Carlone15icraws}
L.~Carlone, J.~Dong, S.~Fenu, G.~Rains, and F.~Dellaert, ``Towards 4{D} crop
  analysis in precision agriculture: Estimating plant height and crown radius
  over time via expectation-maximization,'' in {\em ICRA Workshop on Robotics
  in Agriculture}, 2015.

\end{thebibliography}

\end{document}